\documentclass[11pt]{article}
\usepackage{xspace}

\usepackage[final]{acl}

\usepackage{times}
\usepackage{latexsym}

\usepackage[T1]{fontenc}

\usepackage[utf8]{inputenc}

\usepackage{microtype}

\usepackage{inconsolata}

\usepackage{graphicx}
\newcommand{\paratitle}[1]{\vspace{1.5ex}\noindent\textbf{#1}}
\newcommand{\ie}{\emph{i.e.,}\xspace}

\newcommand{\ignore}[1]{}
\usepackage{amsmath} 
\usepackage{amsfonts}
\usepackage{booktabs} 
\usepackage{subcaption} 
\usepackage{enumitem}
\usepackage{graphicx}
\usepackage{subcaption}
\usepackage{multirow} 
\usepackage{bm}
\usepackage{booktabs}
\usepackage{caption}
\usepackage{algorithm}
\usepackage{algpseudocode}
\usepackage{float}
\title{Beyond Fully Random Masking: Attention-Guided Denoising and Optimization for Diffusion Language Models}

\author{
Jia Deng$^{1,5}$, Junyi Li$^{2}$\thanks{Corresponding author}, Wayne Xin Zhao$^{1,5}$\footnotemark[1], Jinpeng Wang$^{3}$, Hongyu Lu$^{4}$, Ji-Rong Wen$^{1,5}$ \\
$^{1}$Gaoling School of Artificial Intelligence, Renmin University of China \\
$^{2}$Department of Data Science, City University of Hong Kong \\
$^{3}$Meituan \quad $^{4}$WeChat, Tencent \\
$^{5}$Beijing Key Laboratory of Research on Large Models and Intelligent Governance \\
\texttt{dengjia0510@outlook.com, junyili@cityu.edu.hk,  batmanfly@gmail.com} \\
}

\begin{document}
\maketitle
\begin{abstract}
Diffusion large language models (dLLMs) offer an efficient alternative to autoregressive models through parallel decoding, yet existing post-training methods largely rely on random masking strategies that overlook intrinsic token dependencies. In this work, we present an empirical analysis of attention in dLLMs and show that tokens attending more strongly to unmasked context exhibit greater generation stability and play a critical role in reasoning.
Motivated by these findings, we propose \textbf{AGDO}, an attention-guided denoising and optimization framework that aligns both training and optimization with attention-derived dependencies. AGDO determines the denoising order based on attention structure and emphasizes attention-critical tokens during supervised fine-tuning and reinforcement learning. Experiments on mathematical and coding benchmarks demonstrate that AGDO consistently improves reasoning performance, outperforming state-of-the-art post-training methods for dLLMs.
\end{abstract}

\section{Introduction}
Diffusion large language models (dLLMs) have recently emerged as a promising alternative to autoregressive (AR) models for language modeling~\citep{li2025survey}. Unlike AR models, which generate tokens sequentially from left to right, dLLMs iteratively denoise and decode tokens in parallel, offering substantial efficiency advantages during inference~\citep{khannAGDO025mercury,gong2025diffucoder}. Recent models such as LLaDA~\citep{nie2025large,zhu2025llada} and Dream~\citep{ye2025dream} have demonstrated performance competitive with state-of-the-art AR models~\citep{llama3modelcard,team2024qwen2} of comparable scale, highlighting the growing potential of diffusion-based language modeling.

Despite these advances, effectively post-training dLLMs remains challenging. Existing post-training approaches for full-attention dLLMs, including diff-GRPO~\citep{zhao2025d1} and wd1~\citep{tang2025wd1}, typically rely on randomly masking tokens and optimizing the model over the masked positions. While simple and efficient, such strategies fail to align with the actual inference dynamics of dLLMs, leading to a mismatch between training and inference. Several recent works attempt to reduce this discrepancy by introducing different masking strategies. Blockwise supervised fine-tuning~\citep{sun2025blockwise} adopts a semi-autoregressive unmasking order, while other approaches~\citep{wang2025revolutionizing} follow the natural left-to-right generation order during training. Although these methods improve training stability and efficiency, they still impose externally defined decoding orders and overlook a critical property of full-attention dLLMs: under bidirectional attention, token dependencies are not strictly determined by positional order, but emerge dynamically through attention interactions.


To better understand these intrinsic dependencies, we conduct an empirical analysis of attention patterns in dLLMs. Our analysis reveals two key observations. First, attention distributions exhibit strong sparsity and temporal consistency across denoising steps, indicating that each token consistently relies on a small and stable set of context tokens. Second, tokens that attend more heavily to already denoised tokens exhibit significantly higher probability stability during generation, suggesting that the denoising order plays a crucial role in maintaining generation reliability. These findings imply that an effective training strategy for dLLMs should explicitly align the denoising trajectory with attention-induced token dependencies.

Motivated by these insights, we propose \textbf{AGDO} (\textbf{A}ttention-\textbf{G}uided \textbf{D}enoising and \textbf{O}ptimization), a two-stage post-training framework that explicitly aligns the denoising trajectory of dLLMs with intrinsic attention structure. AGDO first derives an attention-guided denoising order from valid attention scores, which govern the token sequence to be unmasked during training. By ensuring that tokens are denoised only after sufficient attention-supported context is available, this denoising order forms the foundation of AGDO.
On top of this method, AGDO performs supervised fine-tuning (\textbf{AGDO-SFT}) and reinforcement learning (\textbf{AGDO-RL}), while further emphasizing attention-hub tokens through an attention-based re-weighting strategy.


Extensive experiments on challenging mathematical and coding benchmarks demonstrate that AGDO significantly improves the reasoning capabilities of masked dLLMs, consistently outperforming existing post-training methods. These results validate the importance of aligning training dynamics with intrinsic attention dependencies and highlight attention-guided denoising as a principled design choice for dLLMs.


\begin{figure*}[!t]
  \centering
  \begin{subfigure}[t]{0.24\linewidth}
    \includegraphics[width=\linewidth]{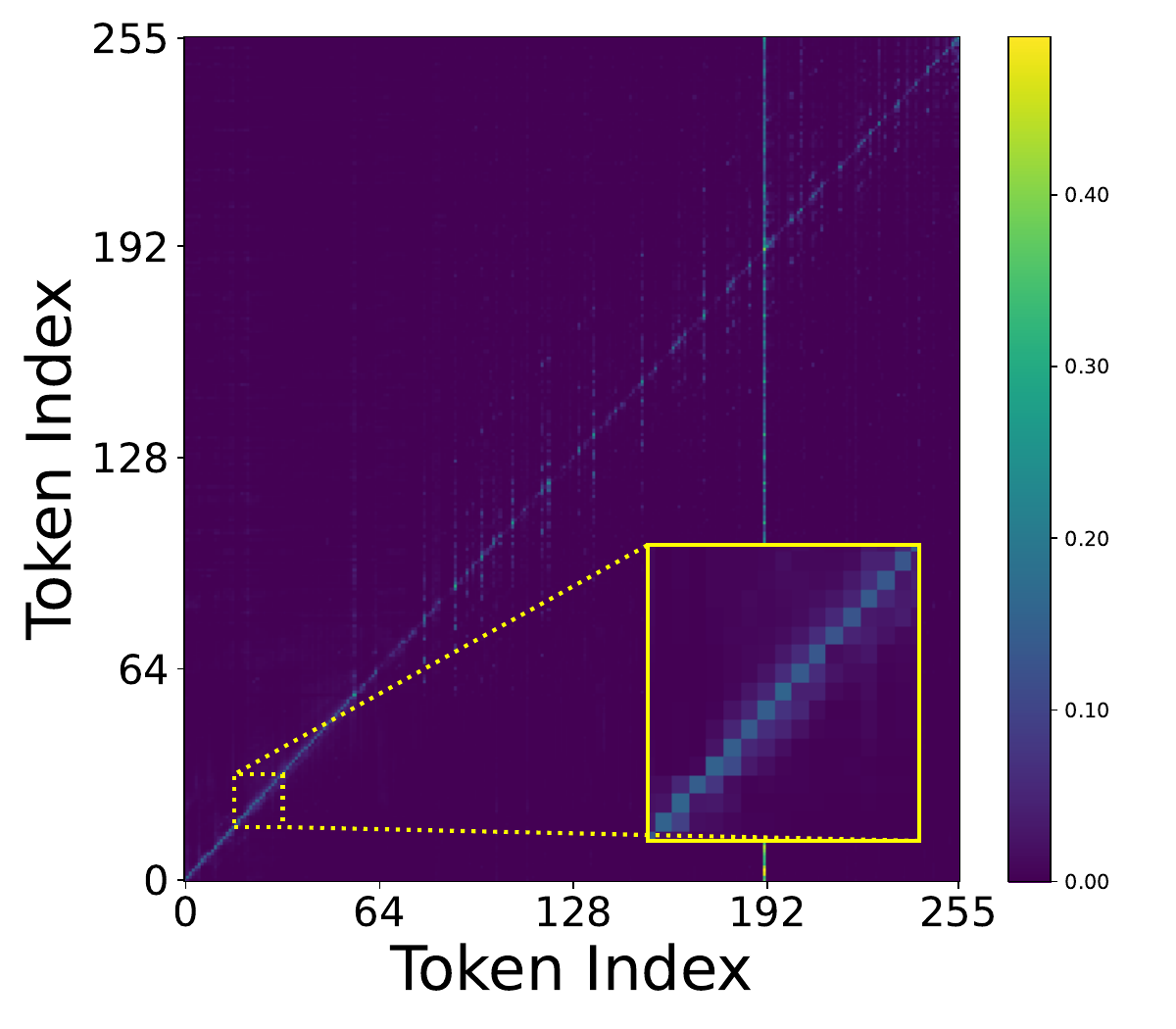}
    \caption{Step 29.}
    \label{fig:dream_matrix1}
  \end{subfigure}\vspace{0.1em}
  \begin{subfigure}[t]{0.24\linewidth}
    \includegraphics[width=\linewidth]{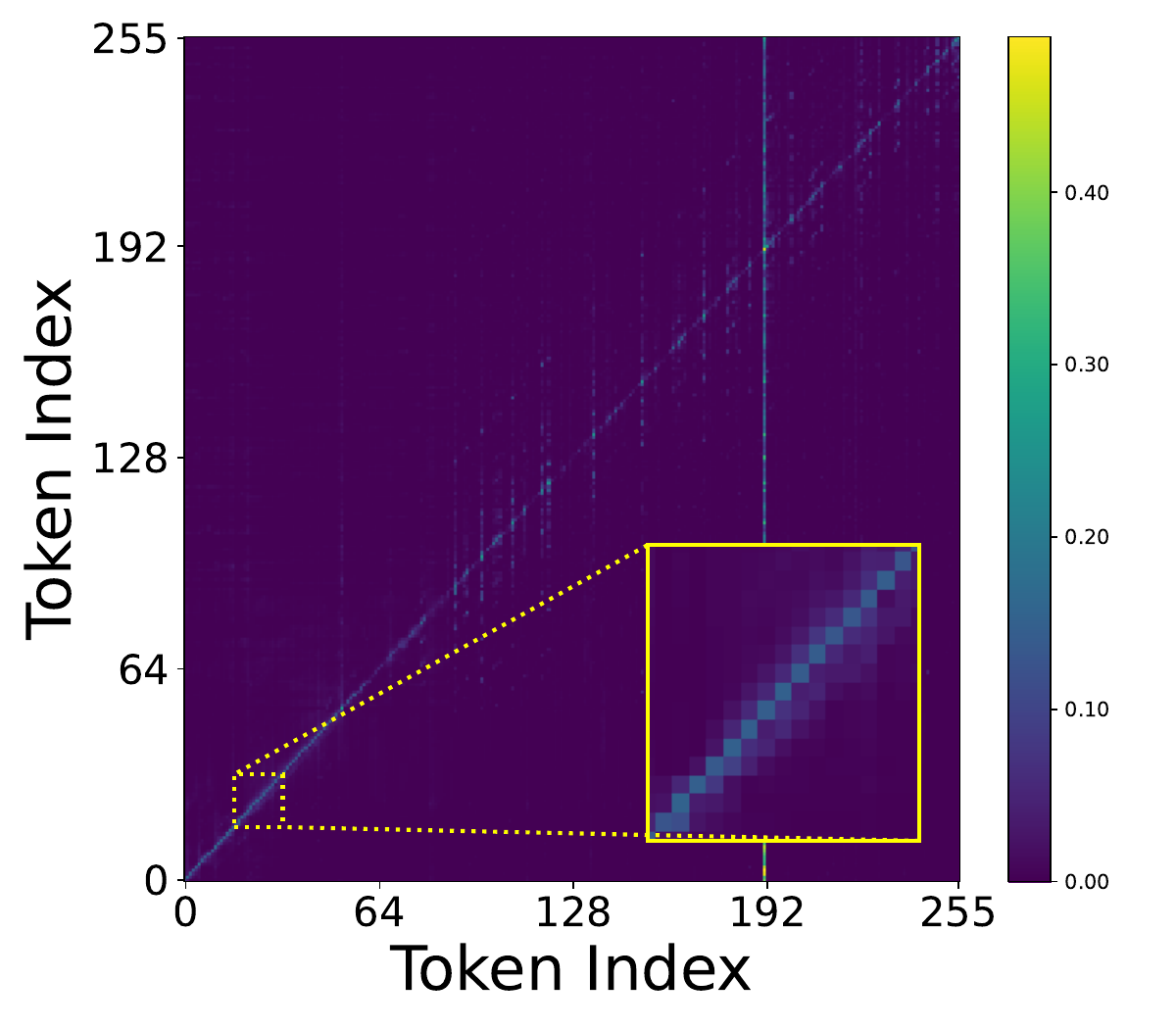}
    \caption{Step 35.}
    \label{fig:dream_matrix2}
  \end{subfigure}\vspace{0.1em}
  \begin{subfigure}[t]{0.24\linewidth}
    \includegraphics[width=\linewidth]{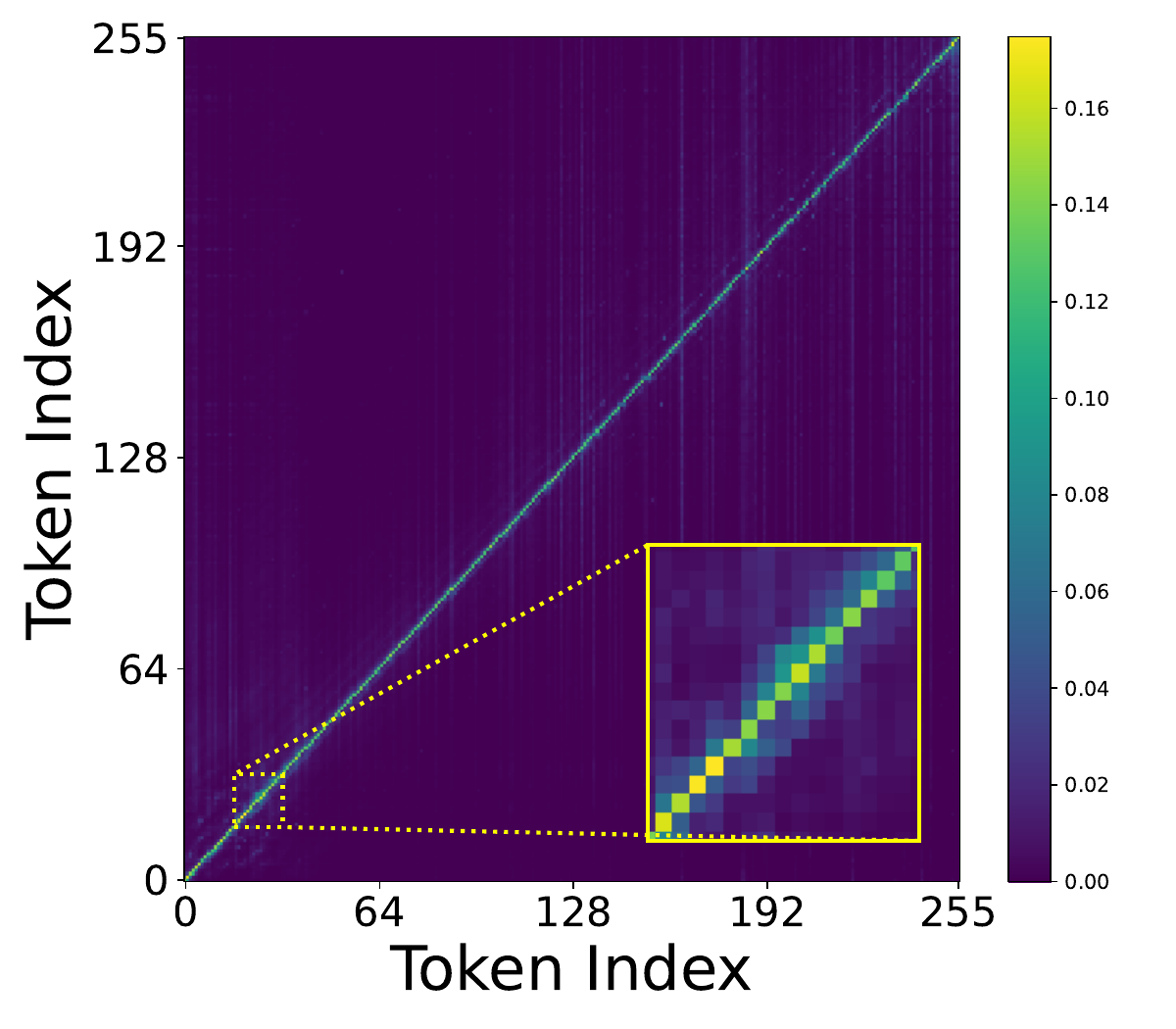}
    \caption{Step 15.}
    \label{fig:dream-token1}
  \end{subfigure}\vspace{0.1em}
  \begin{subfigure}[t]{0.24\linewidth}
    \includegraphics[width=\linewidth]{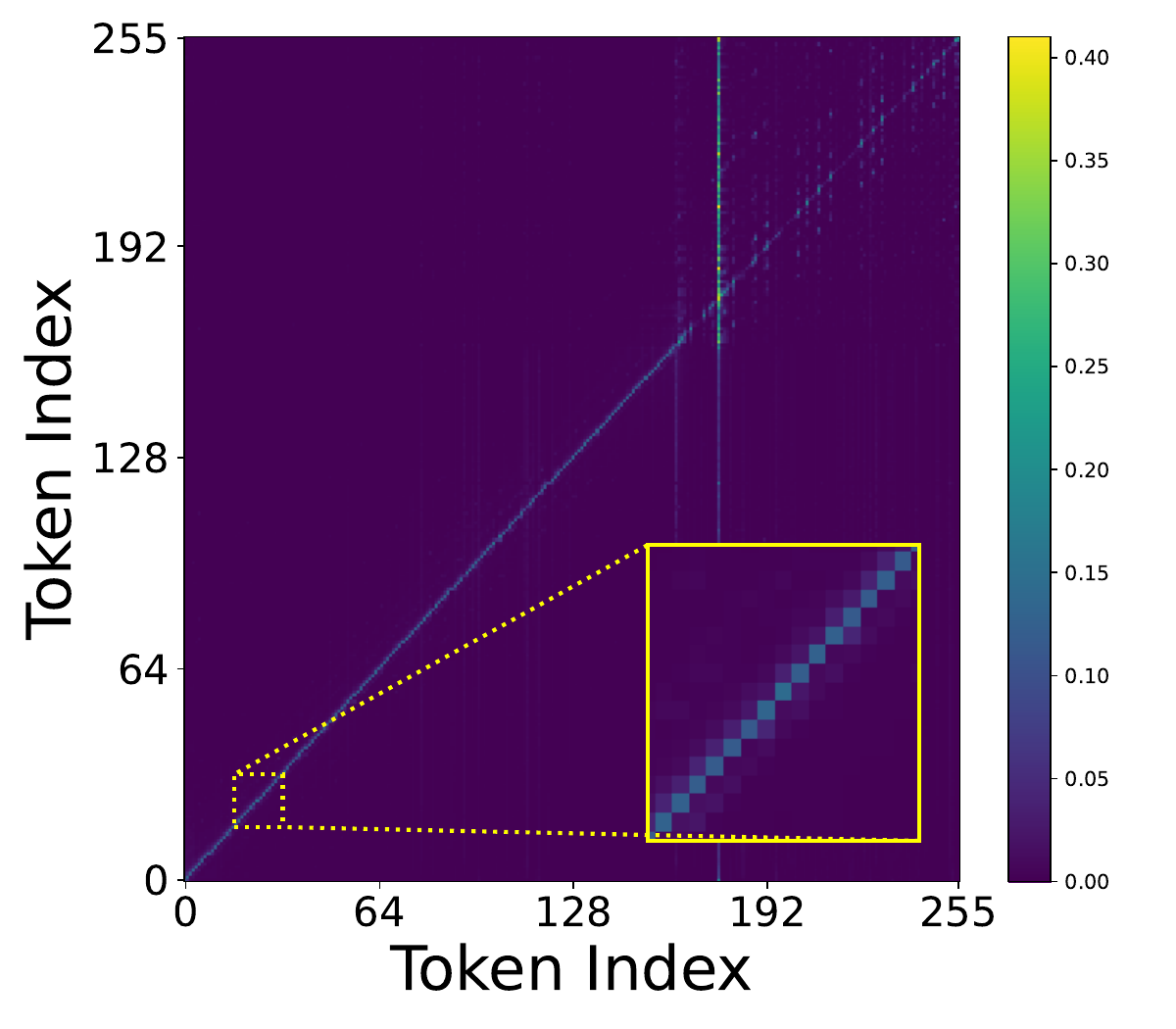}
    \caption{Step 60.}
    \label{fig:dream-token2}
  \end{subfigure}
  
  \vspace{0.1em} 
  
  \begin{subfigure}[t]{0.24\linewidth}
    \includegraphics[width=\linewidth]{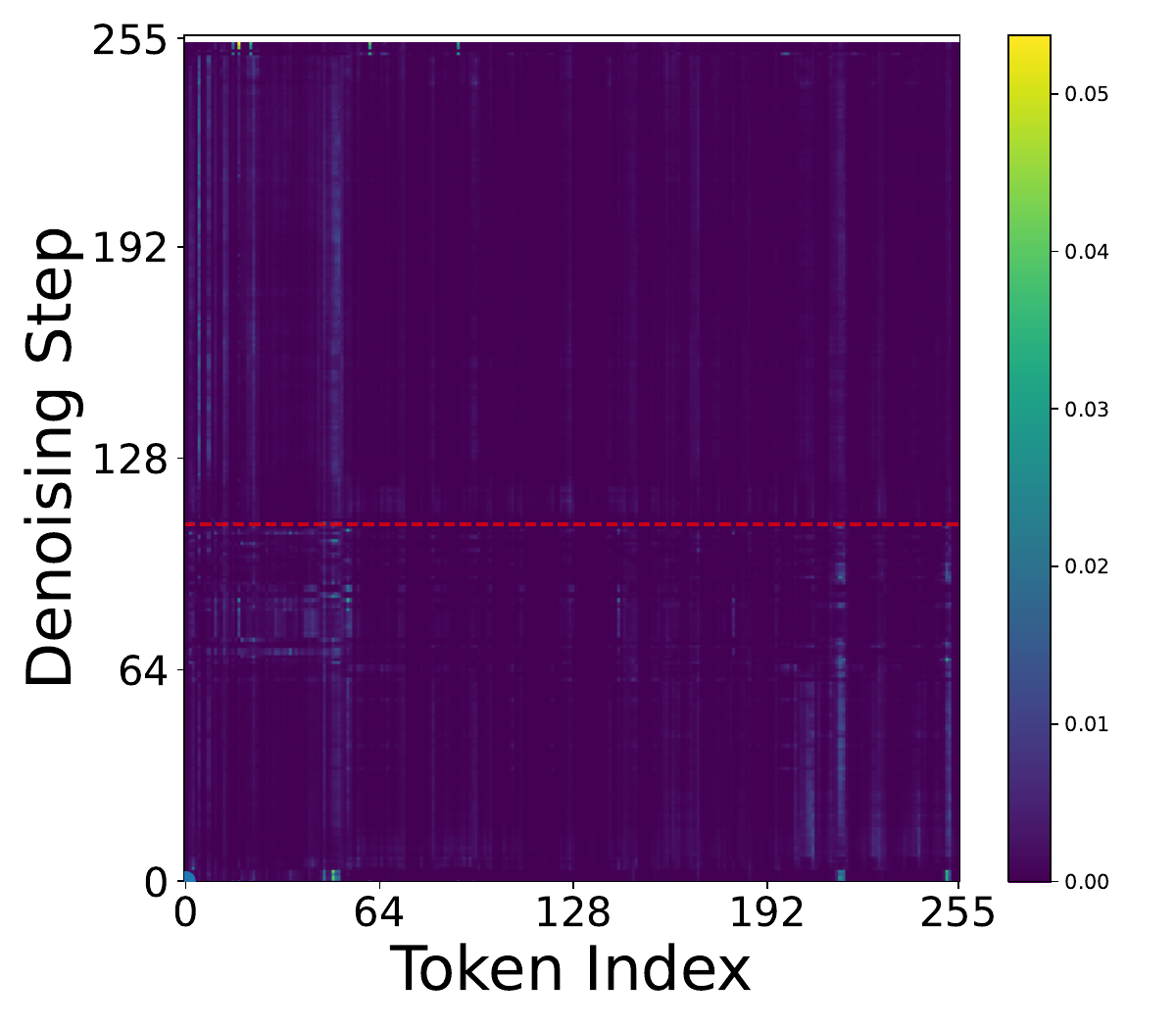} 
    \caption{Token 61.}
    \label{fig:img_e}
  \end{subfigure}\vspace{0.1em}
  \begin{subfigure}[t]{0.24\linewidth}
    \includegraphics[width=\linewidth]{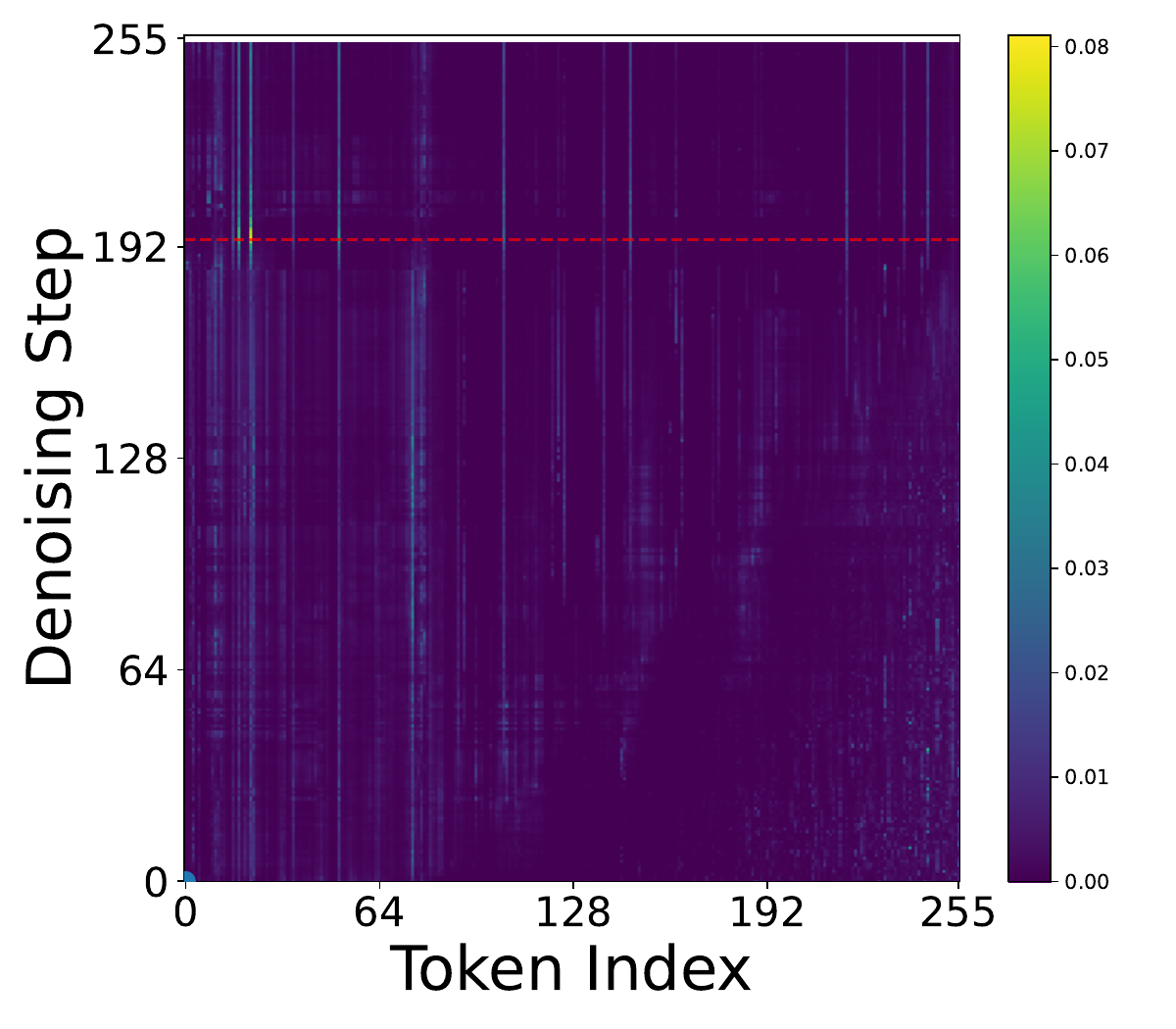} 
    \caption{Token 195.}
    \label{fig:img_f}
  \end{subfigure}\vspace{0.1em}
  \begin{subfigure}[t]{0.24\linewidth}
    \includegraphics[width=\linewidth]{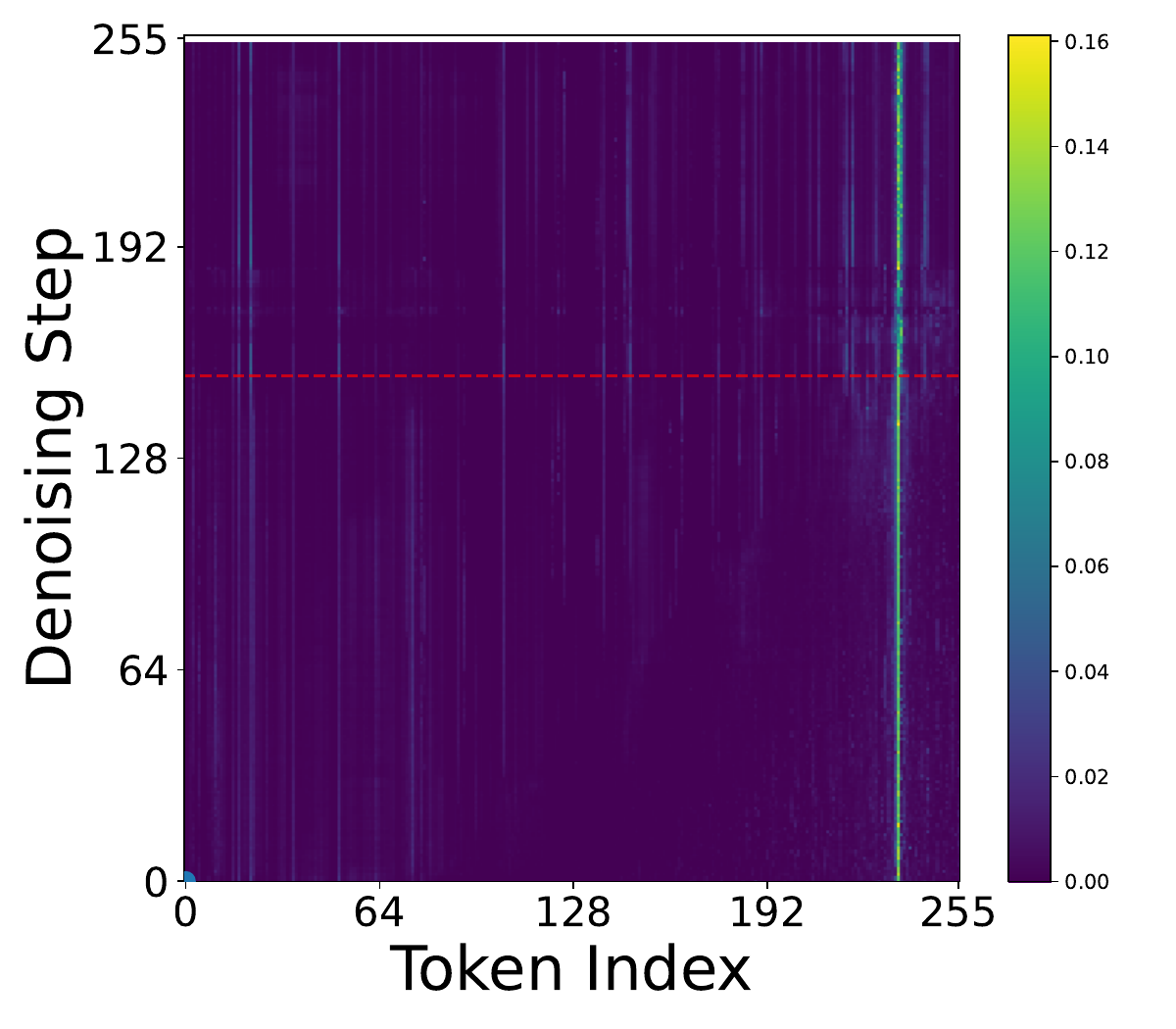} 
    \caption{Token 154.}
    \label{fig:img_g}
  \end{subfigure}
    \vspace{0.1em}
  \begin{subfigure}[t]{0.24\linewidth}
    \includegraphics[width=\linewidth]{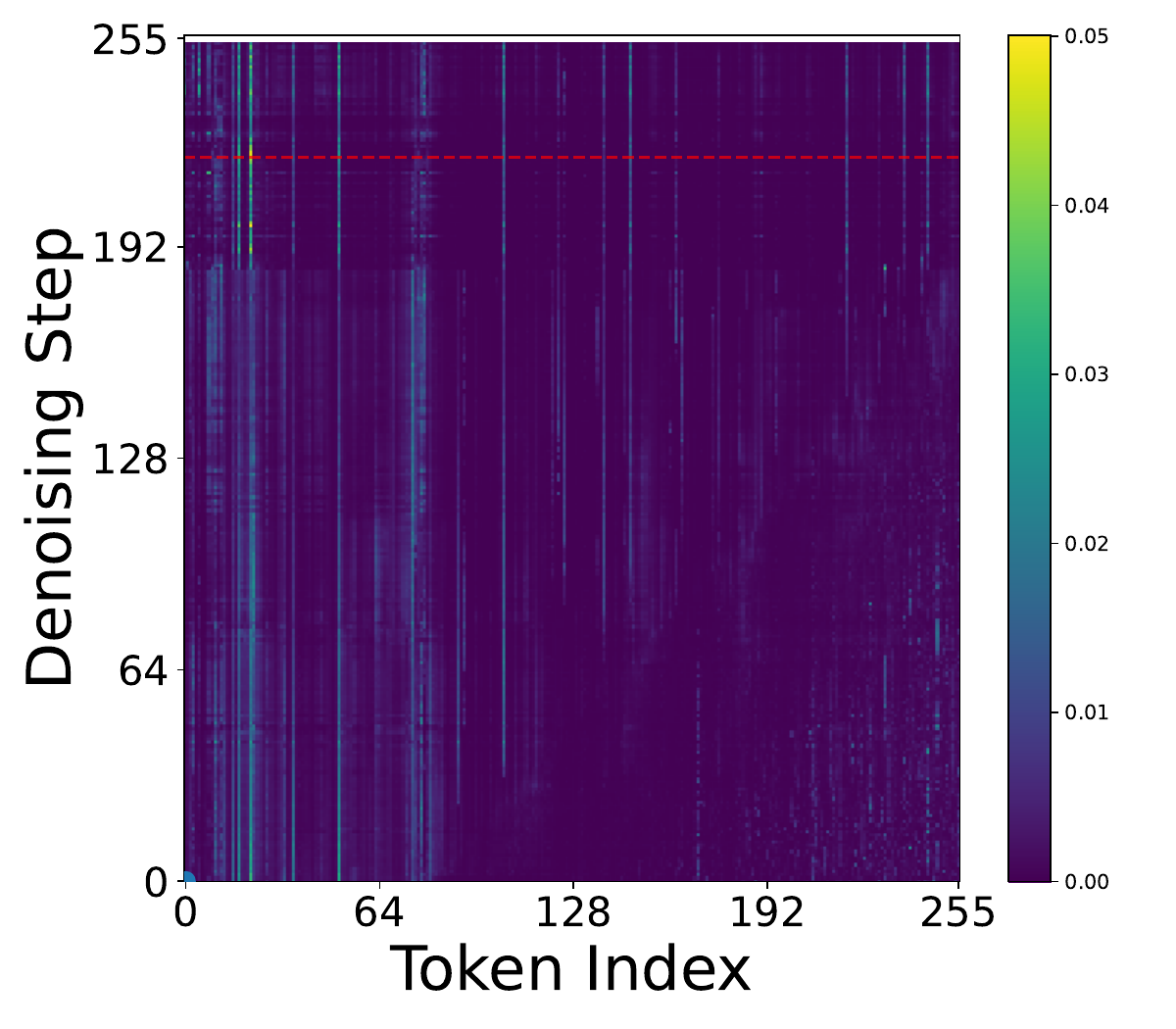} 
    \caption{Token 220.}
    \label{fig:img_h}
  \end{subfigure}
  \caption{Attention dynamics during the denoising process on Dream-v0-Instruct-7B~\citep{ye2025dream}. Figure~(a) to (d) display the attention distributions among tokens at randomly selected denoising steps. Figure~(e) to (h) illustrate how the attention of a randomly chosen token towards other tokens changes throughout the denoising process. The red dashed lines indicate the steps at which the corresponding token is unmasked.}
  \label{fig:dream-all-attention}
\end{figure*}


\section{Related Work}\label{sec:related_work}
\paratitle{Diffusion Large Language Models.}
DLLMs recently attract attention as an alternative to AR models~\citep{li2025survey}. Early attempts to apply diffusion to text employ continuous latent representations~\citep{gong2022diffuseq,li2022diffusion}. Masked diffusion, where masked tokens are iteratively predicted, proves more scalable~\citep{ye2025dream,nie2025large,cheng2025sdar}. Recent models such as DiffuLlama~\citep{gong2024scaling} and Dream~\citep{ye2025dream} adapt pre-trained language models with masked diffusion objectives. In contrast, LLaDA~\citep{nie2025large} shows that training large diffusion-based language models from scratch using full-attention achieves performance comparable to AR models like Llama~\citep{dubey2024Llama}. Most recent work focuses on accelerating inference for dLLMs, introducing strategies such as parallel decoding~\citep{gao2025self}, speculative decoding~\citep{cheng2025deer} and KV-caching~\citep{wu2025fast,liu2025dllm}. While previous work primarily focuses on pre-training and inference acceleration for dLLMs, we focus on post-training methods to enhance their reasoning capabilities.

\paratitle{DLLMs Reasoning Advancement.}
Current approaches to enhancing reasoning in dLLMs mainly center around post-training and inference stages. Inference-time methods such as remasking-based optimization~\citep{li2025adaptive,bao2025learning} and variable-length adaptation~\citep{denoisingbeyond} enable the modeling of longer reasoning chains and utilize the bidirectional context of diffusion models. In the post-training stage, reinforcement learning methods such as group-based advantage estimation and preference optimization are explored~\citep{wang2025spg,tang2025wd1,gong2025diffucoder}. Additionally, diffu-GRPO~\citep{zhao2025d1} applies mean-field approximation to improve computational efficiency. LLaDA 1.5~\citep{zhu2025llada} addresses variance in Evidence Lower Bound~(ELBO) based likelihood estimation by employing preference optimization. However, these approaches often overlook the subtle dependencies among tokens that complex reasoning tasks require. Instead, our approach explicitly incorporates attention-derived signals to identify and leverage token dependencies during post-training.

\section{Preliminaries}
\subsection{DLLMs with Full Attention}
Unlike autoregressive models that employ causal masking, dLLMs~\citep{nie2025large,ye2025dream} employ bidirectional self-attention, allowing each token to attend to the entire sequence. Specifically, for the $l$-th layer and the $h$-th attention head, the attention score matrix $A^{(l,h)}$ is computed as:
\begin{equation}
A^{(l,h)} = \mathrm{softmax}\left( \frac{Q^{(l,h)} K^{(l,h)\top}}{\sqrt{d_k}} \right),
\label{equ:attn_score}
\end{equation}
where no attention mask is used, thereby enabling full contextual visibility across all positions.

During the forward process, dLLMs progressively corrupt the original sequence $x_0$ into a noisy sequence $x_t$, with the time $t \in [0,1]$ controlling the corruption level. Each token is independently masked according to the transition distribution:
\begin{equation}
q_{t|0}(x_t^i \mid x_0^i) = 
\begin{cases} 
\beta_t, & x_t^i = x_0^i \\
1-\beta_t, & x_t^i = \texttt{[MASK]}
\end{cases},
\label{eq:masking}
\end{equation}
where $\beta_t$ decreases with $t$, and $\beta_1 = 0$ at $t = 1$, indicating the sequence is fully masked. The training objective is to reconstruct the masked tokens conditioned on the partially corrupted sequence $x_t$. Let $\mathcal{M}_t$ denote the set of token indices that are masked at time step $t$. The model is trained by minimizing a weighted negative log-likelihood, corresponding to the evidence lower bound (ELBO):
\begin{equation}
\mathcal{L}_{\text{SFT}} = \mathbb{E}_{t, x_0, x_t}
\left[
\frac{1}{t} \sum_{k \in \mathcal{M}_t} \log f_\theta(x_0^k | x_t)
\right].
\label{eq:nelbo}
\end{equation}
\subsection{Policy Optimization for DLLMs}

Reinforcement learning~(RL) for language models is commonly implemented using Proximal Policy Optimization (PPO)~\citep{schulman2017proximal}, which optimizes a policy $\pi_\theta$ by maximizing the objective:
\begin{align}
\small
&\mathbb{E}_{q,o} \Bigg[
\sum_{t=1}^{|o|}
\min\Big( r_t \hat{A}_t,\; 
\text{clip}(r_t,\; 1\!-\!\epsilon,\; 1\!+\!\epsilon) \hat{A}_t \Big) \notag \\
&- \beta \cdot \mathrm{KL}\big[\pi_\theta(\cdot \mid q, o_{<t}) \,\|\, \pi_{\text{ref}}(\cdot \mid q, o_{<t})\big]
\Bigg],
\end{align}
where $r_t$ is the probability ratio between the current policy $\pi_\theta$ and old policy $\pi_{old}$, and $\hat{A}_t$ is the estimated advantage.

To reduce the complexity of PPO, recent work~\citep{shao2024deepseekmath} proposed Group Relative Policy Optimization (GRPO), which eliminates the need for a learned value function by estimating advantages via group-wise normalization. Specifically, given a group of $G$ sampled responses $\{o^1, \dots, o^G\}$ with corresponding rewards $\{R^1, \dots, R^G\}$, the advantage assigned to the $i$-th response is computed as:
\begin{equation}
\small
\hat{A}_t^i = \frac{R^i - \mathrm{mean}(\{R^j\}_{j=1}^G)}{\mathrm{std}(\{R^j\}_{j=1}^G)}.
\end{equation}
However, applying GRPO to dLLMs presents non-trivial challenges. Unlike AR models~\citep{yang2025qwen3,dubey2024Llama}, dLLMs do not admit a sequential factorization over tokens, which complicates the computation of token-level likelihood ratios and KL regularization terms required for policy optimization. Recent approaches address this by adopting mean-field approximations, enabling efficient single-pass estimation of both importance weights and KL divergence for policy optimization~\citep{zhao2025d1,tang2025wd1}.


\section{Attention Analysis in DLLMs}
\label{attention_analysis}
In this section, we conduct an empirical analysis of the attention mechanism in dLLMs to uncover intrinsic token dependencies during the reasoning process. For clarity, we focus on the final layer, denoted as layer $L$, where semantic dependencies are typically most pronounced. Our primary analysis is performed on Dream-v0-Instruct-7B. To demonstrate the generality of our findings, we further validate the results on LLaDA in Appendix~\ref{app_llada}.

\subsection{Attention Patterns in DLLMs}
\label{attention_pattern}

To gain deeper insights into attention distributions in dLLMs, we utilize queries from the MATH-500 benchmark~\citep{hendrycks2020measuring} and trace the temporal evolution of attention score maps (as defined in Equation~\ref{equ:attn_score}) in the final layer across the denoising trajectory. We average the attention scores across all heads and visulize them in Figure~\ref{fig:dream-all-attention}.

\paratitle{Horizontal sparsity across steps.}
Figures~\ref{fig:dream-all-attention} (a)-(d) show that attention weights in dLLMs exhibit a consistently sparse structure throughout the denoising process. In particular, most tokens primarily attend to themselves and their immediate neighbors, forming prominent diagonal patterns similar to those observed in AR models~\citep{xiao2023efficient,hsieh2024found}. In addition, we observe distinct vertical structures manifested as bright columns, indicating that a small subset of key tokens attracts concentrated attention from many other positions. Notably, these sparse patterns remain stable across different denoising steps, corroborating prior observations in recent work~\citep{song2025sparse}.

\paratitle{Vertical consistency across steps.} Figures~\ref{fig:dream-all-attention} (e)-(h) further reveal strong temporal consistency in token-wise attention patterns across denoising steps. Although the query, key, and value representations are recomputed at each inference step, a given token attends to largely the same set of tokens. This behavior appears as persistent vertical lines in the attention maps, suggesting that the most salient contextual dependencies for each token are largely invariant to the masking state. This finding is again consistent with the analysis in~\citet{song2025sparse}.

\subsection{Impact of Valid Attention}

The observations in Section~\ref{attention_pattern} motivate us to examine how the denoising level of the attended tokens influences the current token. To quantify this effect, we define the \emph{valid attention score} of token $i$ at its denoising step as:
\begin{equation}
    S_i = \sum_{k \in \mathcal{U}} \left( \frac{1}{H} \sum_{h=1}^{H} A^{(L, h)}_{i, k} \right),
    \label{eq:valid_attn}
\end{equation}
where $A^{(L, h)}_{i, k}$ denotes the normalized attention score from token $i$ to token $k$ in the layer $L$ and head $h$, and $\mathcal{U}$ denotes the set of already unmasked tokens.
Furthermore, we track the probability evolution of each token from its initial denoising step until the end of the inference process. We define the probability change $\Delta P_i$ for token $i$ as:
\begin{equation}
    \Delta P_i = P_{i}^{\min} - P_{i}^{\text{denoise}},
    \label{eq:prob_change}
\end{equation}
where $P_{i}^{\text{denoise}}$ is the generation probability of token $i$ at its denoising step, and $P_{i}^{\min}$ is the minimum probability for token $i$ observed during the subsequent inference process.

\begin{figure}[!h]
  \centering
  \includegraphics[width=0.9\linewidth]{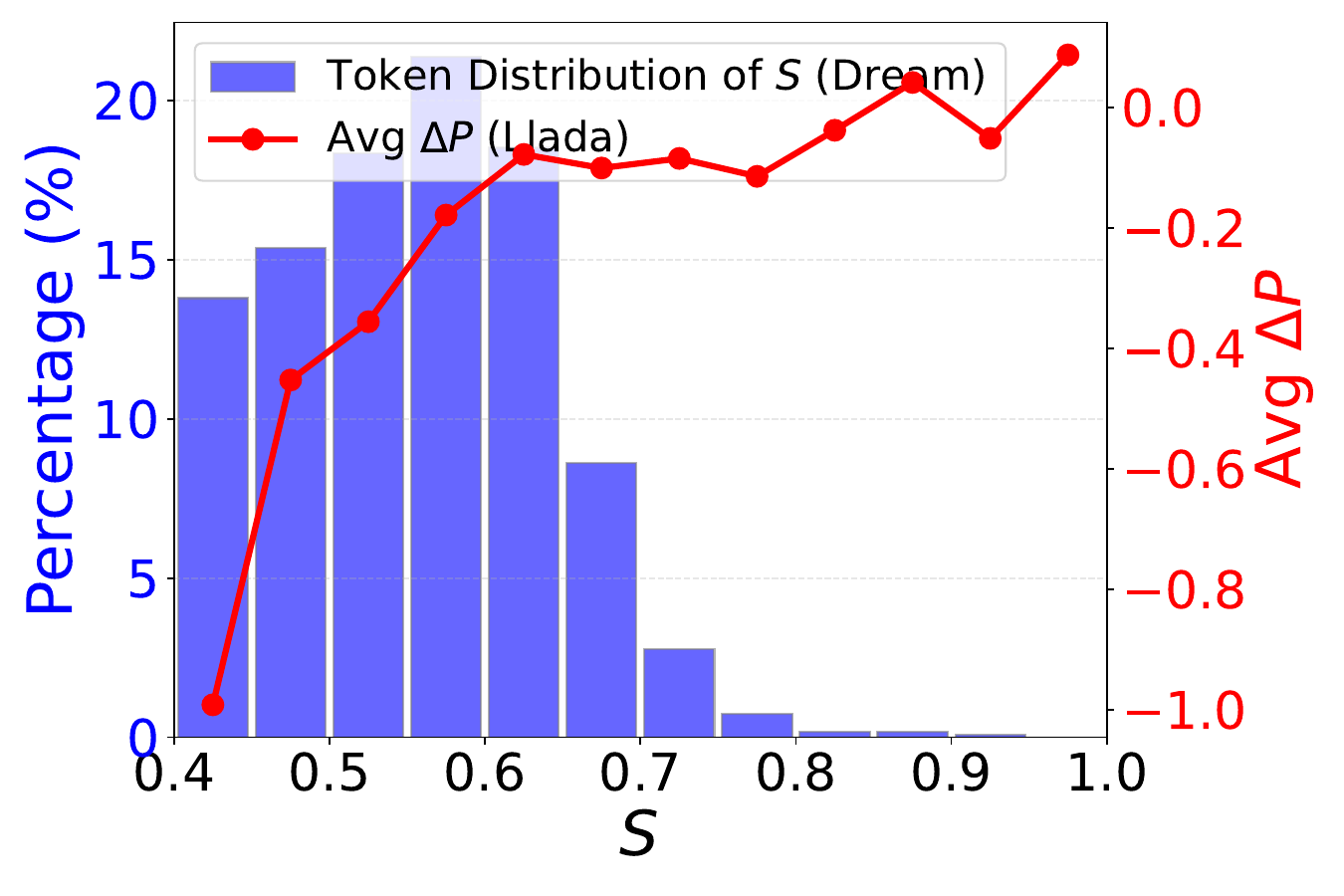}
  \caption{Relationship between the valid attention score $S$ and the average probability change $\Delta P$.}
  \label{fig:cor_results}
\end{figure}

Figure~\ref{fig:cor_results} illustrates the distribution of valid attention scores $S$ and the corresponding average probability change $\Delta P$. We observe a clear positive correlation between $S$ and $\Delta P$. Specifically, tokens that allocate more attention to already unmasked tokens (i.e., higher $S_i$) exhibit greater stability, in the sense that their probabilities are less likely to decrease as subsequent tokens are generated.

To further explore the influence of $S$ on generation quality, we introduce a hybrid token selection strategy that augments the probability-based baseline with the valid attention score $S$. At each denoising step, we first construct a candidate set consisting of masked tokens whose probabilities exceed 0.9. Among these candidates, we select the token with the highest $S$ value to be unmasked.

As shown in Table~\ref{tab:accuracy_comparison}, this $S$-aware denoising strategy consistently outperforms the probability-only baseline across different predefined sequence lengths on MATH-500. These results indicate that incorporating valid attention information leads to a denoising order that better captures intrinsic token dependencies, thereby producing more reliable and higher-quality generations.


\begin{table}[!h]
    \centering
    \resizebox{0.8\linewidth}{!}{ 
        \setlength{\tabcolsep}{12pt}
        \begin{tabular}{lccc}
            \toprule
            \textbf{Strategy} & \textbf{128} & \textbf{256} & \textbf{512} \\
            \midrule
            Max-Prob & 16.8 & 13.8 & 13.0 \\ 
            Max-$S$    & 21.2 & 16.2 & 15.2 \\
            \bottomrule
        \end{tabular}
    }
    \caption{Accuracy on MATH-500 under static sampling with different predefined sequence lengths. Max-Prob selects the token with the highest probability, while Max-$S$ selects the token with the highest valid attention score among high-probability candidates.}
    \label{tab:accuracy_comparison}
\end{table}

\section{Approach}

Motivated by our attention analysis in Section~\ref{attention_analysis}, which reveals strong structural sparsity, temporal consistency, and stability-aware token dependencies in dLLMs, we propose \textbf{AGDO}, an \emph{Attention-Guided Denoising and Optimization} framework. AGDO explicitly aligns training and optimization with intrinsic attention dependencies.
Under a shared attention-guided denoising order, AGDO supports two training regimes: 
\textbf{AGDO-SFT}, which performs attention-guided supervised fine-tuning, and 
\textbf{AGDO-RL}, which applies attention-guided denoising optimization.

\subsection{Attention-Guided Denoising Order}
\label{decode_order}

Our analysis in Section~\ref{attention_pattern} shows that tokens allocating more attention to already unmasked tokens exhibit significantly higher probability stability during the denoising process. This observation suggests that the denoising order in dLLMs should respect attention-induced token dependencies.

To operationalize this insight, we construct an attention-guided denoising order based on the valid attention score $S$ defined in Equation~\ref{eq:valid_attn}. For a complete generation trajectory, we perform a single forward pass at the final denoising timestep to obtain the attention score matrix in the last layer. The set of unmasked tokens $\mathcal{U}$ is initialized with all prompt tokens. At each denoising step $t$, we select the top-$n$ tokens with the highest $S$ scores and add them to $\mathcal{U}$. In the next step, valid attention scores are recomputed for the remaining tokens based on the updated $\mathcal{U}$. This procedure is repeated until all tokens are assigned to a denoising step.

By denoising tokens after sufficient attention-supported context is available, this ordering aligns the generation trajectory with intrinsic dependency structures captured by attention. The detailed algorithm is provided in Appendix~\ref{app:AGDOtg}.


\subsection{Attention-Guided Fine-Tuning}

Existing supervised fine-tuning (SFT) methods for dLLMs typically rely on either random masking or fixed semi-autoregressive masking strategies~\citep{zhao2025d1,sun2025blockwise} to compute cross-entropy loss on a preselected subset of tokens. Yet, such approaches ignore the attention-induced dependency structure revealed by our analysis. 

In contrast, {AGDO-SFT} follows the attention-guided denoising order described in Section~\ref{decode_order}. At each timestep $t$, we randomly mask only the tokens assigned to that denoising step, ensuring that training conditions mirror the intended inference trajectory. Moreover, inspired by findings on attention centrality in AR models~\citep{lin2024critical,li2025attention}, we further distinguish tokens by their influence on the rest of the sequence. For each token $k$, we define an \emph{influence score} $I_k$ as the total attention it receives from other tokens:
\begin{equation}
    I_k = \sum_{i} \left( \frac{1}{H} \sum_{h=1}^{H} A^{(L, h)}_{i, k} \right),
    \label{equ:influence}
\end{equation}
where $A^{(L,h)}_{i,k}$ denotes the attention weight from token $i$ to token $k$ in the final layer $L$.


Tokens with higher $I_k$ function as attention hubs and exert disproportionate influence on the generation of other tokens. To reflect this, we weight the cross-entropy loss at denoising step $t$ as:
\begin{equation}
\resizebox{0.86\linewidth}{!}{$
    -\mathbb{E}_{t, x_0, x_t}
    \left[
    \frac{1}{|\mathcal{U}_t|} \sum_{k \in \mathcal{U}_t} (1+\gamma_k I_k)\log f_\theta(x_0^k \mid x_t)
    \right].
$}
\label{eq:attn_received_score}
\end{equation}

This design aligns both the masking schedule and the loss weighting with attention structure, enabling the model to focus on tokens that are most critical for global consistency. The full procedure is described in Appendix~\ref{app:AGDOft}.


\subsection{Attention-Guided Policy Optimization}

We further extend AGDO to reinforcement learning under the GRPO algorithm. As in AGDO-SFT, denoising follows the attention-guided order defined in Section~\ref{decode_order}. Furthermore, to emphasize tokens that garner greater attention during the generation process, we augment the advantage estimation by incorporating $I$ derived in Equation~\ref{equ:influence}.
The resulting AGDO-RL objective is formulated as:
\begin{equation}
\resizebox{0.87\linewidth}{!}{$ 
\begin{aligned}[b]
&-\mathbb{E}_{t, q, o} \Bigg[
    \frac{1}{|\mathcal{U}_t|} \sum_{k \in \mathcal{U}_t}
    \min\Big( r_k \hat{A}_k',\; 
        \text{clip}(r_k,\; 1\!-\!\epsilon,\; 1\!+\!\epsilon) \hat{A}_k' \Big) \\
&\quad - \beta\, D_{\mathrm{KL}}\left( \phi_{\pi_\theta}(\cdot \mid q)\,\|\,\phi_{\pi_{\mathrm{ref}}}(\cdot \mid q) \right)
\Bigg],
\end{aligned}
$}
\label{equ:grpo_dllm}
\end{equation}
where the attention-adjusted advantage $\hat{A}_k'$ is defined as:
\begin{equation}
\hat{A}_k' = \hat{A}_k + sign(\hat{A}_k) \cdot \delta \cdot I_k,
\label{equ:adv_new}
\end{equation}
where $\delta$ controls the strength of attention guidance. By amplifying policy updates for tokens that are central in the attention graph, AGDO-RL aligns preference optimization with the intrinsic reasoning structure of dLLMs. Implementation details are provided in Appendix~\ref{app:AGDOpo}.

\begin{table*}[h!]
\centering
\footnotesize 
\renewcommand{\arraystretch}{1.0}
\setlength{\tabcolsep}{3pt} 
\scalebox{0.98}{ 
\begin{tabular}{l c c c c c c}
\toprule
        & \multicolumn{3}{c}{\textbf{Math}} 
        & \multicolumn{2}{c}{\textbf{Code}} 
        & \multirow{2}{*}{\textbf{Average}} \\
        \cmidrule(lr){2-4} \cmidrule(lr){5-6}
        & \textbf{GSM8K} & \textbf{MATH500} & \textbf{Minerva} 
        & \textbf{LiveBench} & \textbf{LiveCodeBench-v2} 
        & \\
\midrule
\multicolumn{7}{c}{\textit{Similar-sized AR LLMs}} \\
\midrule
Llama3.1-8B-Instruct~\citep{llama3modelcard}                 & 84.5 & 51.9 & 37.5   & 19.7   & 20.0   & 42.7   \\
Qwen2.5-7B-Instruct~\citep{team2024qwen2}                        & 89.9 & 74.0 & 50.4   & 31.1   & 26.9   & 54.5   \\
\midrule
\multicolumn{7}{c}{\textit{Masked DLLMs baselines}} \\
\midrule
Dream-v0-Instruct-7B~\citep{ye2025dream}                        & 69.4 & 38.9 & 11.6   & 10.7   & 10.7   &  28.3  \\
LLaDA-8B-Instruct~\citep{nie2025large} & 81.5 & 38.3 & 11.9& 4.9 & 5.9 & 28.5 \\
\midrule
\multicolumn{7}{c}{\textit{SFT from Dream}} \\
\midrule
SFT    & 83.5 & 48.3 & \underline{14.8} & \underline{11.3} & 11.5 & 33.9 \\
blockwise SFT     & \textbf{86.0} & \underline{51.7} & 12.3 & 10.2 & \underline{11.8} & \underline{34.4} \\
\textbf{AGDO-SFT} (ours)     & \underline{85.3} & \textbf{53.7} & \textbf{15.3} & \textbf{12.5} & \textbf{13.1} & \textbf{36.0} \\
\midrule
\multicolumn{7}{c}{\textit{RL from Dream}} \\
\midrule
Diff-GRPO~\citep{zhao2025d1}& 85.0 & 45.5 & 15.3 & 15.2 & 13.9 & 35.0 \\
Coupled RL~\citep{gong2025diffucoder}& 86.1 & 48.8 & 14.1 & 13.8 & 11.8 & 34.9 \\
TraceRL~\citep{wang2025revolutionizing}& 86.3 & 52.8 & \underline{16.4} & 14.0 & 13.0 & 36.5     \\
\textbf{AGDO-RL} (ours) & \textbf{87.7} & \underline{53.7} & 16.1 & \underline{18.3} & \underline{14.7} & \underline{38.1} \\
 \textbf{AGDO}(ours) & \underline{86.9} & \textbf{56.2} & \textbf{17.0} & \textbf{18.4} & \textbf{15.6} & \textbf{38.8} \\
\bottomrule
\end{tabular}
}
\caption{The main benchmark results across different math and coding tasks on Dream-v0-Instruct-7B. Best results are in \textbf{bold}, and second best are \underline{underlined}.}
\label{tab:main_results}
\end{table*}
\section{Experiment}



\subsection{Experimental Setup}

\paratitle{Models and Benchmarks.}
We mainly conduct our experiments on Dream-v0-Instruct-7B~\citep{ye2025dream}, and further assess the generalization of our approach on LLaDA-8B-Instruct~\citep{nie2025large}. We evaluate model performance across two representative domains: mathematical reasoning and code generation. For mathematical reasoning, we benchmark on MATH-500~\citep{hendrycks2020measuring}, GSM8K~\citep{cobbe2021training}, and Minerva~\citep{lewkowycz2022solving}, covering problems of varying difficulty and reasoning depth. For code generation, we report results on LiveBench~\citep{white2024livebench} and LiveCodeBench-V2~\citep{jain2024livecodebench}, which provide dynamic and execution-based evaluation of real-world coding tasks.


\paratitle{Baselines and Metrics.}
We compare our approach against several established baselines under both SFT and RL. For SFT, we consider two baselines: (i) \textbf{standard SFT} with a fully random masking strategy, and (ii) \textbf{blockwise SFT}~\citep{sun2025blockwise}. To ensure a fair comparison, we augment the training data for the fully random masking baseline by applying 35 independent random masks per example, resulting in an average of approximately 39 forward passes per data point, which is comparable to that of other methods.
For RL, we benchmark against \textbf{Diff-GRPO}~\citep{zhao2025d1}, \textbf{Couple-RL}~\citep{gong2025diffucoder}, and \textbf{TraceRL}~\citep{wang2025revolutionizing}. Similarly, to maintain fairness across methods, we augment the training data for Diff-GRPO and Couple-RL with 15 independent random masks, aligning the average number of forward passes per data point with our approach.
For robust evaluation, each experiment is repeated 8 times, and we report the average accuracy across runs. During inference, we adopt a static decoding strategy with a sampling temperature of 0.1, unmasking one token per denoising step, and a maximum response length of 1,024 tokens.
More implementation details can be found in Appendix~\ref{app_imp}.
\subsection{Main Results}

\label{sec:main}
\begin{figure}[h]
  \centering
  \begin{subfigure}[t]{0.5\linewidth}
    \includegraphics[width=\linewidth]{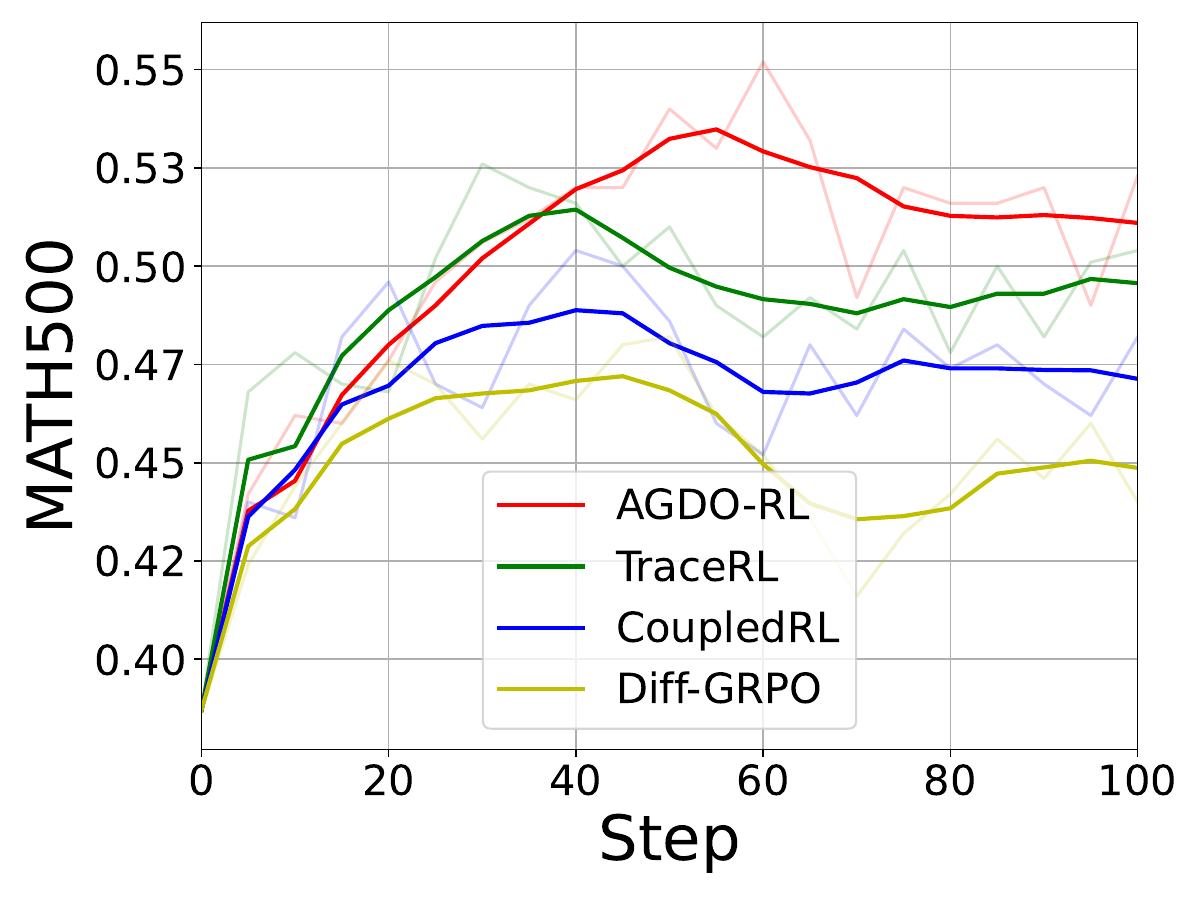}
    \caption{Accuracy on MATH500.}
    \label{fig:test}
  \end{subfigure}\hfill
  \begin{subfigure}[t]{0.5\linewidth}
    \includegraphics[width=\linewidth]{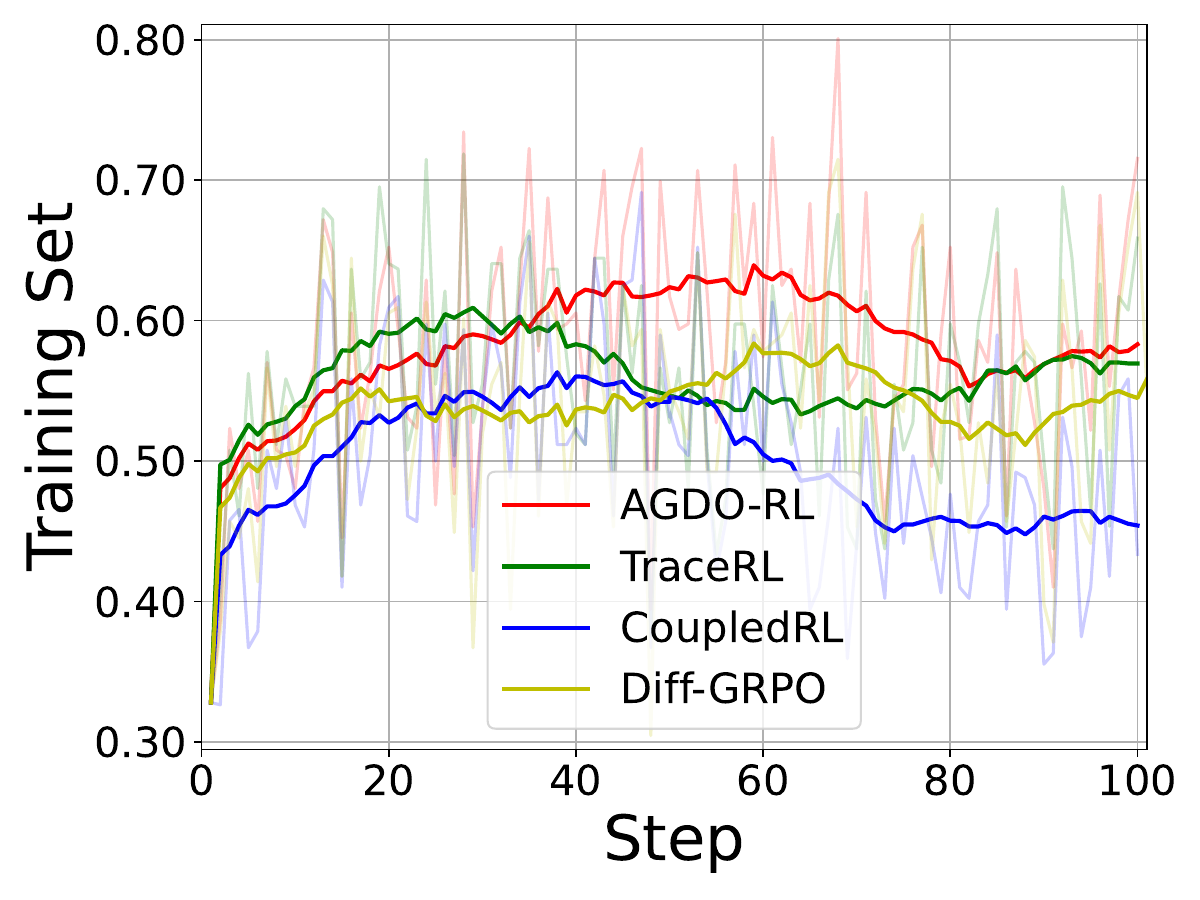}
    \caption{Accuracy on MATH and GSM8K training sets.}
    \label{fig:train}
  \end{subfigure}
  \caption{Accuracy changes on training and testing sets during reinforcement training on Dream-v0-Instruct-7B.}
  \label{fig:dream-results}
\end{figure}

Table~\ref{tab:main_results} presents the quantitative comparison on Dream-v0-Instruct-7B. In the SFT stage, our AGDO-SFT consistently outperforms both standard and blockwise SFT baselines, achieving an average accuracy of 36.0\% compared to 34.4\% for blockwise SFT. Notably, AGDO-SFT yields a significant 3.0\% gain compared to blockwise SFT on the challenging Minerva benchmark, confirming the benefit of aligning training with intrinsic attention dependencies. In the RL stage, AGDO-RL establishes new state-of-the-art results among baselines. Specifically, AGDO-RL achieves 18.3\% on LiveBench, substantially exceeding diff-GRPO (15.2\%). Furthermore, as illustrated in Figure~\ref{fig:dream-results}, our method demonstrates more sustained accuracy growth and superior stability during training on both training and testing sets, indicating that attention-guided optimization effectively mitigates learning difficulties in dLLMs. 
\subsection{Ablation Studies}
\label{sec:abli}
\subsubsection{Ablation on $\gamma$ and $\delta$}
\begin{figure}[H]
  \centering
  \begin{subfigure}[t]{0.5\linewidth}
    \includegraphics[width=\linewidth]{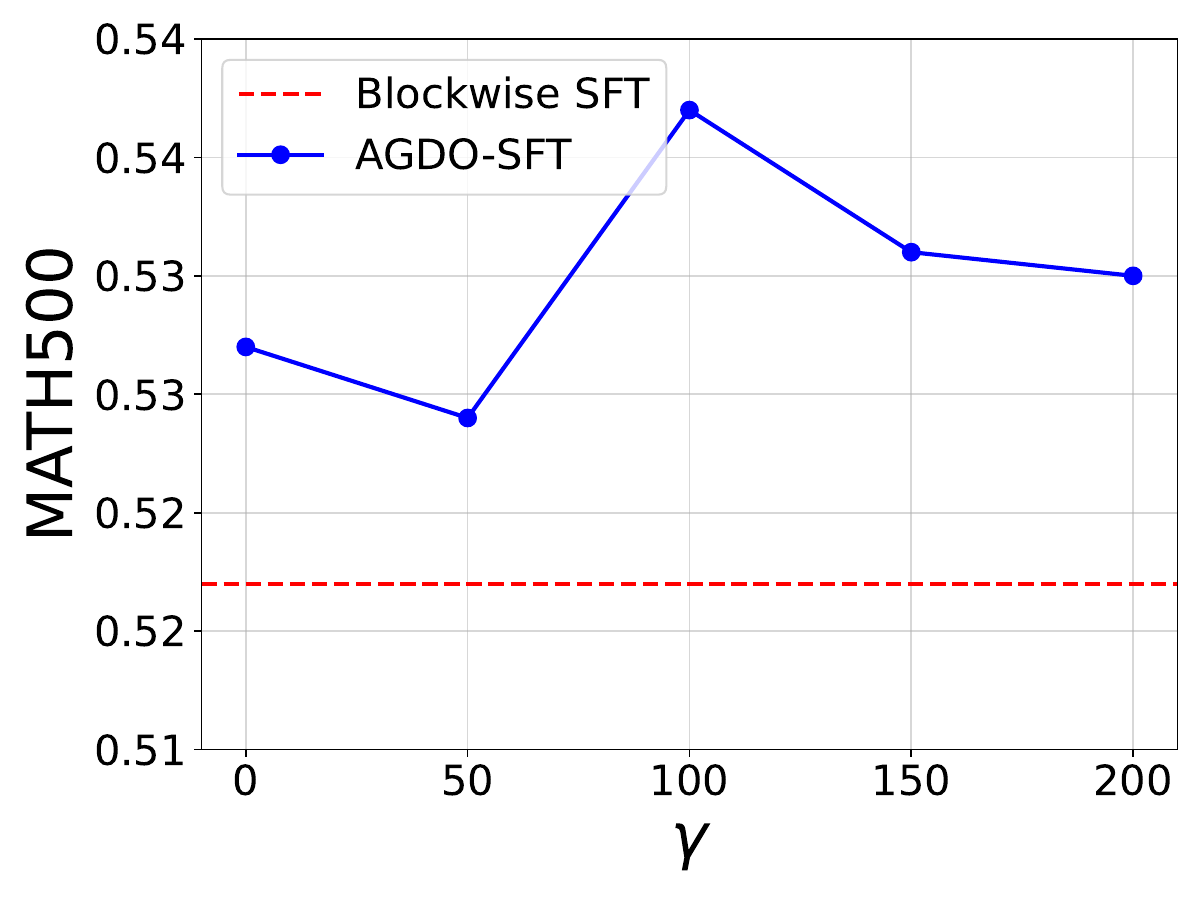}
    \caption{Accuracy changes on MATH500 under different $\gamma$.}
    \label{fig:test}
  \end{subfigure}\hfill
  \begin{subfigure}[t]{0.5\linewidth}
    \includegraphics[width=\linewidth]{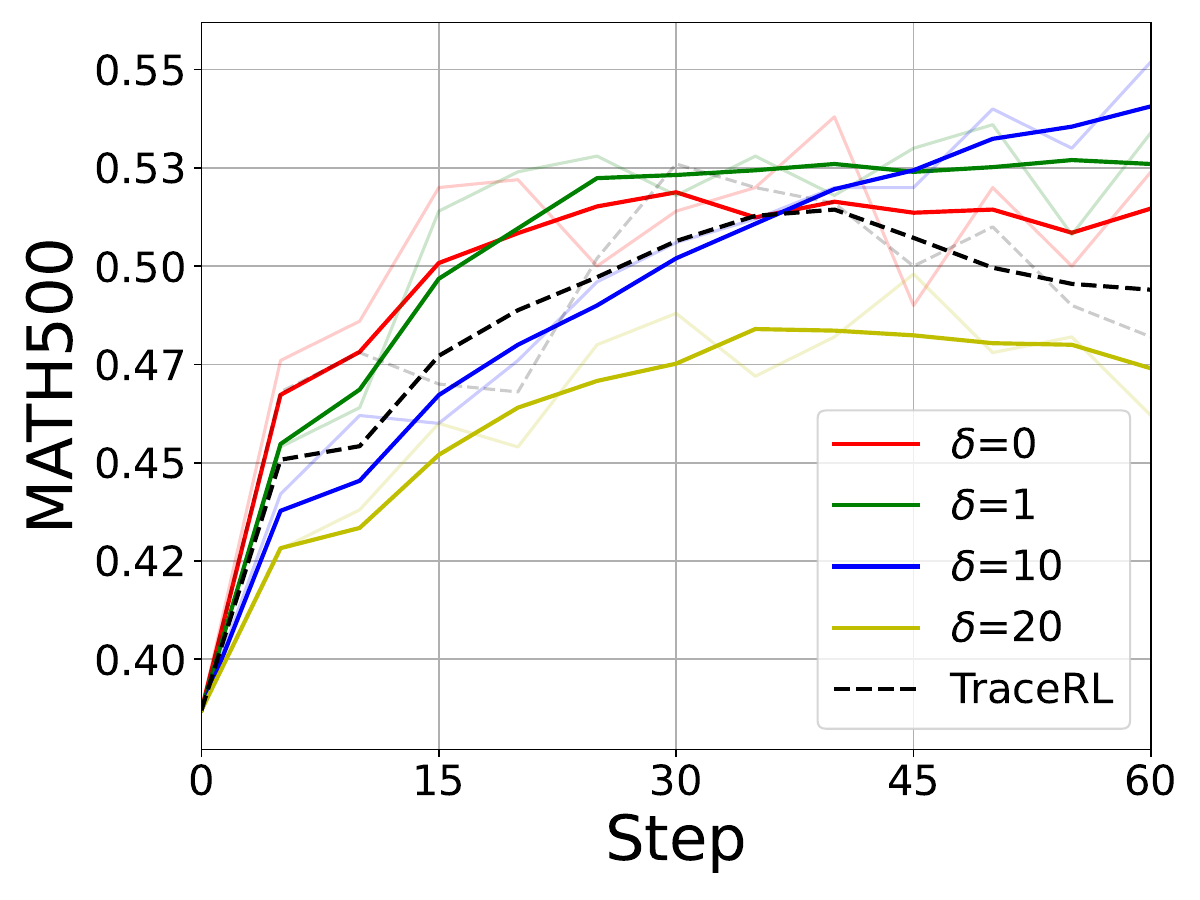}
    \caption{Accuracy changes on MATH500 under different $\delta$.}
    \label{fig:train}
  \end{subfigure}
  \caption{Ablication results on $\gamma$ and $\delta$.}
  \label{fig:abli_dream}
\end{figure}

We investigate the impact of hyperparameters $\gamma$ and $\delta$ on attention-based weighting in AGDO-SFT and AGDO-RL through ablation studies. As illustrated in Figure~\ref{fig:abli_dream}, tuning $\gamma$ during the fine-tuning stage consistently improves accuracy over the blockwise SFT baseline, with $\gamma=100$ yielding the best result on MATH500. Notably, AGDO-SFT outperforms blockwise SFT by approximately 2\% even when $\gamma=0$, a setting that implies the absence of $I$-enhanced training weights. This result demonstrates the isolated effectiveness of aligning the denoising order with attention based dependence. 
In the RL phase, the accuracy curve for $\delta=0$ consistently remains above that of TraceRL, further validating the proposed strategy of aligning the denoising trajectory with attention. We also observe that setting $\delta < 10$ leads to additional gains in training accuracy on top of the attention-guided denoising strategy. Conversely, performance deteriorates when $\delta=20$. We hypothesize that an excessively large $\delta$ induces drastic gradient updates, which violates the trust region constraints fundamental to Proximal Policy Optimization.
\subsubsection{Ablation on Sampling Configs}

\begin{table}[h]
\centering

\footnotesize 

\renewcommand{\arraystretch}{1.2}

\setlength{\tabcolsep}{3pt}

\resizebox{\linewidth}{!}{
    \begin{tabular}{lcccccc}
    \toprule
    \multirow{2}{*}{\textbf{Methods}} & \multicolumn{5}{c}{\textbf{Block Size}} & \multirow{2}{*}{\textbf{Average}} \\
    \cmidrule(lr){2-6}
                            & 8 & 16 & 32 & 64 & 128 & \\
    \midrule
    SFT                  & 48.4 & 48.2 & 48.7 & 49.9 & 50.5 & 49.1 \\
    Blockwise SFT        & 43.9 & 46.3 & 46.3 & 45.5 & 46.9 & 45.8 \\
    \textbf{AGDO-SFT}            & 49.1 & 48.5 & 50.2 & 50.5 & 49.7 & 49.6 \\
    \midrule
    Diff-GRPO            & 43.9 & 43.5 & 45.5 & 46.9 & 45.9 & 45.1 \\
    Coupled RL           & 47.1 & 48.4 & 49.2 & 49.6 & 51.5 & 49.2 \\
    TraceRL              & 47.5 & 50.5 & 50.3 & 52.9 & 51.9 & 50.6 \\
    \textbf{AGDO-RL}               & 48.0 & 50.6 & 53.3 & 53.7 & 52.3 & 51.6 \\
    \textbf{AGDO}            & 51.5 & 52.8 & 53.4 & 53.9 & 53.6 & 53.0 \\
    \bottomrule
    \end{tabular}
}
\caption{Ablation results on MATH500 with different block sizes under a predefined length of 512.}
\label{tab:math500_ablation}
\end{table}
To evaluate the robustness of our method under stricter context constraints, we conduct ablation studies using a reduced sequence length of $L=512$, in contrast to the $L=1024$ setting employed in the main experiments. Table~\ref{tab:math500_ablation} details the performance across block sizes ranging from 8 to 128. In the fine-tuning stage, naive Blockwise SFT exhibits a marked performance decline (averaging 45.8\%) compared to standard SFT (49.1\%). However, AGDO-SFT effectively mitigates this degradation, achieving an average accuracy of 49.6\% and surpassing standard SFT. This result underscores the efficacy of our proposed method. In the subsequent RL stage, AGDO-RL consistently outperforms all RL baselines across varying block sizes. Notably, AGDO-RL attains the highest average accuracy of 51.6\%, peaking at 53.7\% with a block size of 64. AGDO gains the best results on all block sizes, which further validates the effectiveness of aligning training with attention.

\subsubsection{Application on LLaDA}
\begin{figure}[!h]
  \centering
  \begin{subfigure}[t]{0.5\linewidth}
    \includegraphics[width=\linewidth]{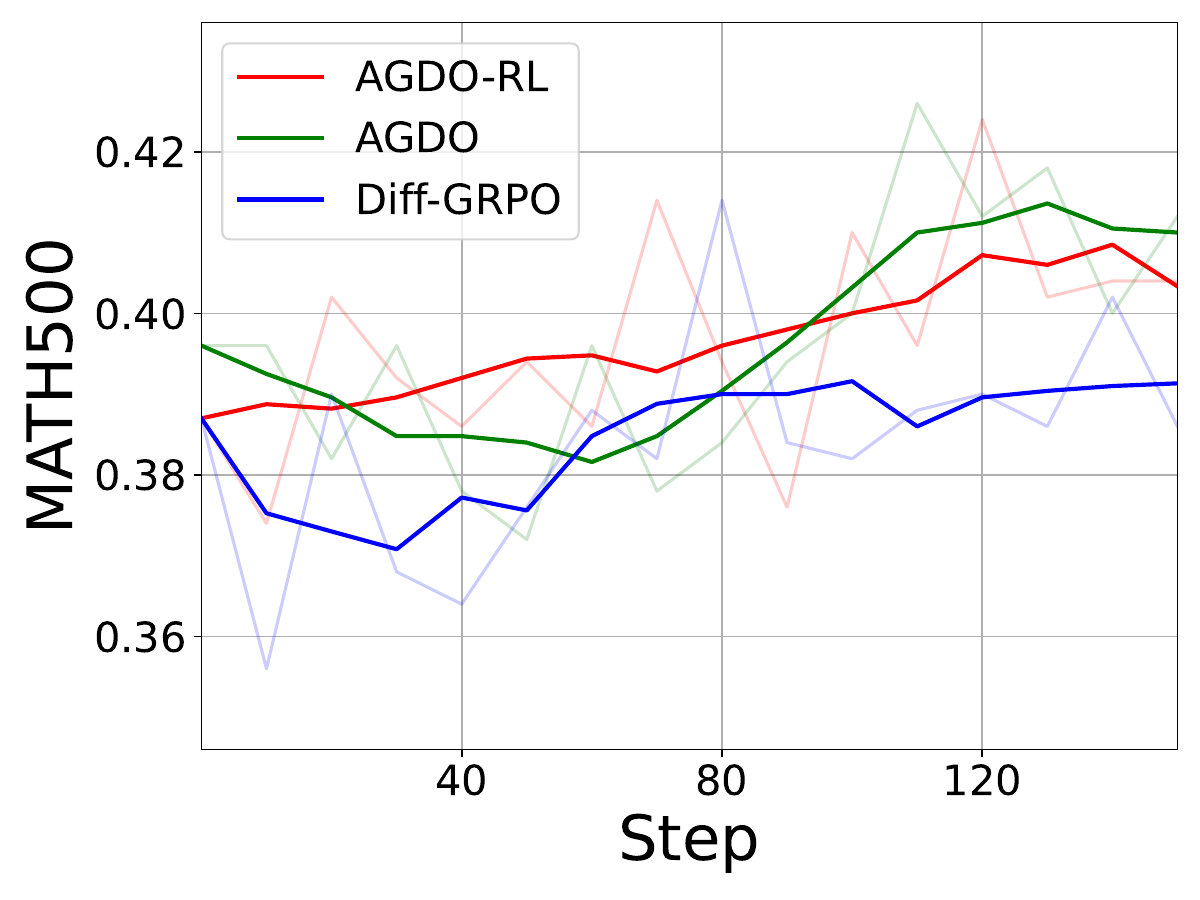}
    \caption{Accuracy on MATH500.}
    \label{fig:test}
  \end{subfigure}\hfill
  \begin{subfigure}[t]{0.5\linewidth}
    \includegraphics[width=\linewidth]{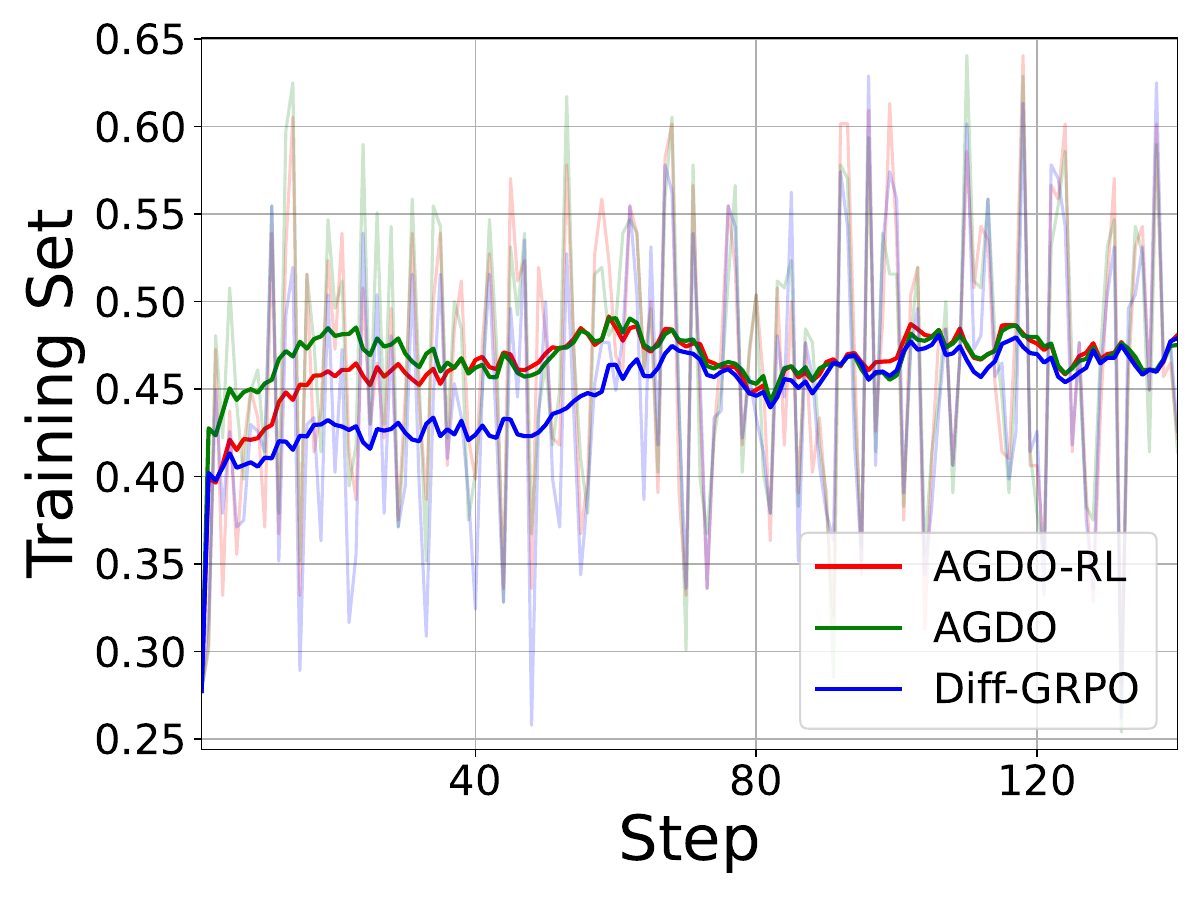}
    \caption{Accuracy on MATH and GSM8K training sets.}
    \label{fig:train}
  \end{subfigure}
  \caption{Accuracy changes during reinforcement training on LLaDA-8B-Instruct~\citep{nie2025large}.}
  \label{fig:llada-results}
\end{figure}

To further validate the effectiveness of our approach, we evaluate our algorithms on LLaDA-8B-Instruct~\citep{nie2025large} using the GSM8K and MATH500 benchmarks, with the training set kept same with Dream. For the SFT stage, we adopt standard SFT with fully random masking and blockwise SFT as baselines. For the full training pipeline, we benchmark against d1-LLaDA~\citep{zhao2025d1}, wd1~\citep{tang2025wd1}, and LLaDA-1.5~\citep{zhu2025llada}. As shown in Table~\ref{tab:llada_results}, our methods outperform the baselines. The accuracy curves in Figure~\ref{fig:llada-results} further show that our methods achieve faster improvement than Diff-GRPO~\citep{zhao2025d1}, demonstrating their effectiveness.

\begin{table}[!h]
\centering
\footnotesize
\renewcommand{\arraystretch}{1.0}
\setlength{\tabcolsep}{7pt}
\begin{tabular}{lcc}
\toprule
\textbf{Methods} & \textbf{GSM8K} & \textbf{MATH500} \\
\midrule
\multicolumn{3}{c}{\textit{Baseline}} \\
\midrule
LLaDA~\citep{nie2025large}          & 81.5 & 38.7 \\
\midrule
\multicolumn{3}{c}{\textit{SFT from LLaDA}} \\
\midrule
SFT             & 82.1 & 38.3 \\
Blockwise SFT     & \underline{82.3} & \underline{38.4} \\
\textbf{AGDO-SFT} (ours)  & \textbf{83.3} & \textbf{39.6} \\
\midrule
\multicolumn{3}{c}{\textit{SFT+RL from LLaDA}} \\
\midrule
d1-LLaDA~\citep{zhao2025d1}         & 82.1 & 40.2 \\
wd1~\citep{tang2025wd1}             & 82.3 & 39.0 \\
LLaDA-1.5~\citep{zhu2025llada}      & \underline{83.3} & \underline{42.6} \\
\textbf{AGDO} (ours)            & \textbf{85.3} & \textbf{42.8} \\
\bottomrule
\end{tabular}
\caption{Results on GSM8K and MATH500 benchmarks for LLaDA-8B-Instruct~\citep{nie2025large}.}
\label{tab:llada_results}
\end{table}
\subsubsection{Ablation on Layer and Head Selection}
\label{sec:ablation_layer_head}

We investigate the impact of layer selection and head aggregation strategy on AGDO-SFT performance. Table~\ref{tab:layer_ablation} compares attention signals extracted from different layers. Using the last layer yields the highest accuracy, consistent with prior findings that deeper layers capture higher-level semantic dependencies~\citep{clark2019does}.

\begin{table}[h]
\centering
\footnotesize
\renewcommand{\arraystretch}{1.1}
\setlength{\tabcolsep}{10pt}
\begin{tabular}{lc}
\toprule
\textbf{Layer Selection} & \textbf{MATH500 Acc (\%)} \\
\midrule
First Layer & 48.8 \\
Middle Layer & 52.5 \\
Last Layer (Ours) & \textbf{53.7} \\
\bottomrule
\end{tabular}
\caption{Ablation on attention layer selection.}
\label{tab:layer_ablation}
\end{table}

Table~\ref{tab:head_ablation} compares different head aggregation strategies. Following~\citet{li2025attention}, we identify local-focused and global-focused heads and compare them against averaging all heads. The all-heads strategy achieves the best result, indicating that both local and global attention patterns contribute to effective denoising guidance.

\begin{table}[h]
\centering
\footnotesize
\renewcommand{\arraystretch}{1.1}
\setlength{\tabcolsep}{10pt}
\begin{tabular}{lc}
\toprule
\textbf{Head Aggregation} & \textbf{MATH500 Acc (\%)} \\
\midrule
Local-focused heads & 53.0 \\
Global-focused heads & 52.2 \\
All heads (Ours) & \textbf{53.7} \\
\bottomrule
\end{tabular}
\caption{Ablation on attention head aggregation for AGDO-SFT.}
\label{tab:head_ablation}
\end{table}

\subsubsection{Ablation on Order vs.\ Weighting}
\label{sec:ablation_order_weight}

To disentangle the contributions of the attention-guided denoising order and the influence-based loss weighting, we conduct a controlled ablation starting from the blockwise SFT baseline. As shown in Table~\ref{tab:order_weight_ablation}, applying the attention-guided order alone (\ie $\gamma=0$) improves accuracy by 1.0\% over blockwise SFT, while adding influence-based weighting alone also yields gains. Combining both components achieves the best performance. Notably, replacing the influence score $I$ with random weights leads to a smaller improvement, confirming that the gains stem from the attention-derived signal rather than stochastic regularization.

\begin{table}[h]
\centering
\footnotesize
\renewcommand{\arraystretch}{1.1}
\setlength{\tabcolsep}{10pt}
\begin{tabular}{lc}
\toprule
\textbf{Method} & \textbf{MATH500 Acc (\%)} \\
\midrule
Blockwise SFT (Baseline) & 51.7 \\
\midrule
Order Only ($\gamma=0$) & 52.7 \\
Weight Only (random order) & 52.4 \\
Random Weight (random order) & 51.9 \\
Order + Weight (Ours) & \textbf{53.7} \\
\bottomrule
\end{tabular}
\caption{Ablation on the individual contributions of the attention-guided denoising order and influence-based weighting in AGDO-SFT.}
\label{tab:order_weight_ablation}
\end{table}
\subsection{Further Analysis}
\subsubsection{Internalization of Attention Alignment}
\label{sec:further}
\begin{figure}[!h]
  \centering
  \includegraphics[width=0.8\linewidth]{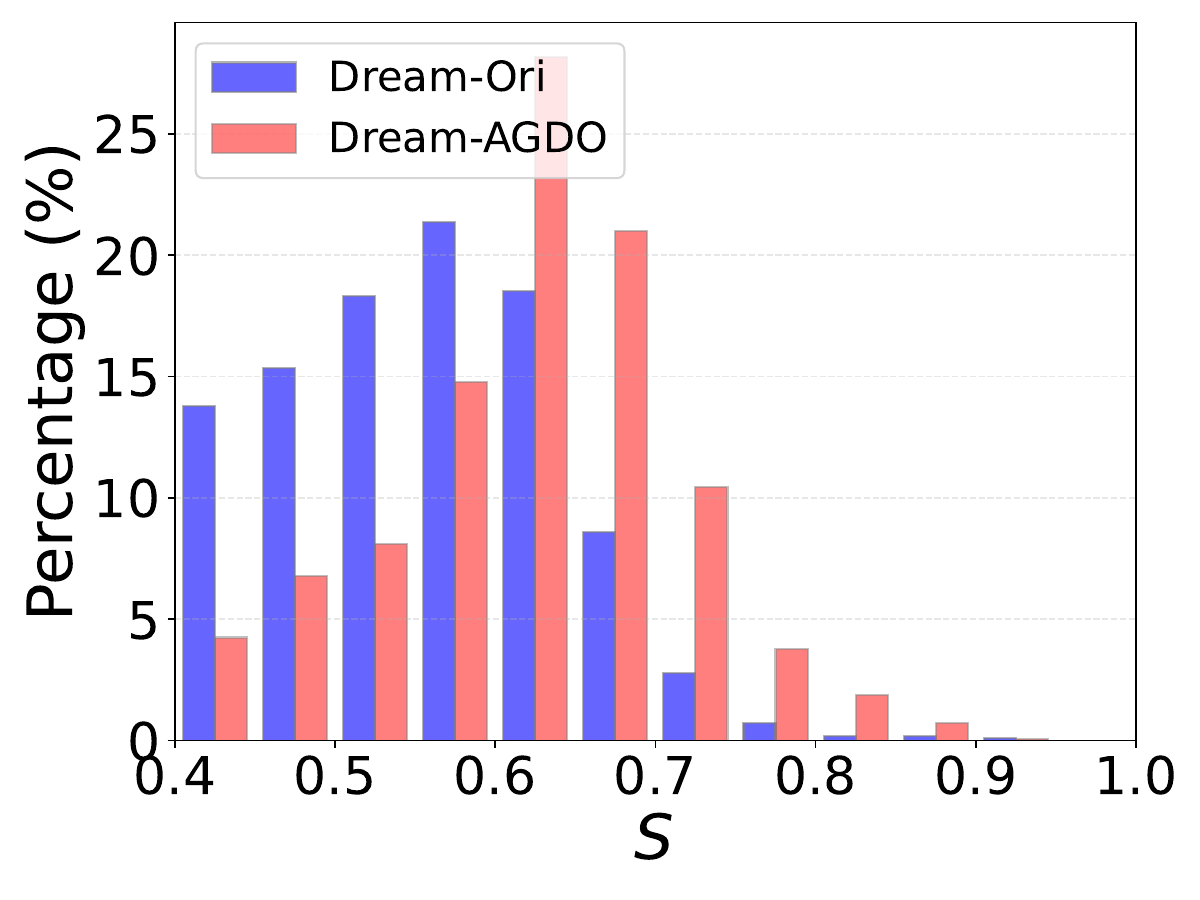}
  \caption{Comparison of average $\Delta P$ and $S$.}
  \label{fig:s_change}
\end{figure}
To investigate the impact of our training framework on internal reasoning mechanisms, we compare the distribution of valid attention score $S$ on the MATH500 benchmark between the original Dream model and the AGDO-trained model. As illustrated in Figure~\ref{fig:s_change}, the distribution of $S$ for the AGDO-trained model exhibits a distinct rightward shift compared to the baseline. This trend indicates that the model has internalized the attention-guided denoising strategy, learning to prioritize the generation of tokens that possess strong dependencies on the already unmasked context. By maximizing valid attention during inference, the model effectively reduces generation uncertainty and establishes more robust logical chains, thereby enhancing the quality of generated responses. This further demonstrates the effectiveness of our method.
\subsubsection{Computational Cost Analysis}
\label{cost_analysis}

A potential concern regarding AGDO-RL is the overhead incurred by the online attention analysis during RL. However, generating a response requires $T$ denoising steps (\ie$T$ forward passes), whereas our method adds only \textbf{one} single forward pass to analyze the full sequence. 

We empirically test the overhead of online attention analysis during RL on Dream-v0-Instruct-7B using 8 NVIDIA H20 GPUs and $T=1024$ denoising steps. Under the same RL settings as the main experiments, Table~\ref{tab:rl_time_breakdown} shows that \textit{Attention Analysis} accounts for only \textbf{3\%} of the total rollout time, indicating that AGDO-RL improves reasoning with negligible impact on training efficiency.

\begin{table}[h]
    \centering
    \small
    \begin{tabular}{lrr}
    \toprule
    \textbf{Stage} & \textbf{Forward Passes} & \textbf{Total Time (s)} \\
    \midrule
    Rollout Stage & 262,144 & 390 \\
    Attention Analysis & 256 & 12 \\
    \midrule
    \textbf{Total} & 262,400 & 402 \\
    \bottomrule
    \end{tabular}
    \caption{Time breakdown of a single RL iteration. The overhead from our Attention Analysis is marginal compared to the Rollout Stage.}
    \label{tab:rl_time_breakdown}
\end{table}
\subsubsection{Generalization to General NLP Tasks}

To verify that AGDO does not introduce regressions on non-reasoning tasks, we evaluate on HellaSwag~\citep{zellers2019hellaswag}  and CommonsenseQA~\citep{talmor2019commonsenseqa} using the same inference configuration. As shown in Table~\ref{tab:general_nlp}, AGDO consistently outperforms all baselines across both SFT and RL stages. Notably, AGDO-SFT surpasses both standard SFT and blockwise SFT on CommonsenseQA by 2.6\% and 5.2\% respectively, confirming that aligning denoising with attention dependencies enhances linguistic coherence and commonsense reasoning without introducing domain-specific bias.

\begin{table}[h]
\centering
\footnotesize
\renewcommand{\arraystretch}{1.1}
\setlength{\tabcolsep}{6pt}
\begin{tabular}{lcc}
\toprule
\textbf{Methods} & \textbf{HellaSwag} & \textbf{CSQA} \\
\midrule
Dream-v0-Instruct-7B & 45.4 & 34.3 \\
\midrule
\multicolumn{3}{c}{\textit{SFT}} \\
\midrule
SFT & 60.9 & 75.5 \\
Blockwise SFT & 57.1 & 72.9 \\
\textbf{AGDO-SFT} (ours) & \textbf{61.8} & \textbf{78.1} \\
\midrule
\multicolumn{3}{c}{\textit{RL}} \\
\midrule
Diff-GRPO & 53.0 & 26.0 \\
Coupled RL & 40.0 & 27.3 \\
TraceRL & 45.7 & 27.6 \\
\textbf{AGDO-RL} (ours) & \underline{58.0} & \underline{35.0} \\
\textbf{AGDO} (ours) & \textbf{64.7} & \textbf{78.3} \\
\bottomrule
\end{tabular}
\caption{Results on general NLP benchmarks. CSQA denotes CommonsenseQA.}
\label{tab:general_nlp}
\end{table}
\section{Conclusion}

In this paper, we revisit post-training for diffusion large language models (dLLMs) and show that existing random masking strategies fail to fully leverage intrinsic token dependencies. To address this limitation, we propose AGDO, a unified framework that integrates attention-guided denoising with supervised and reinforcement learning.
By aligning the denoising order with attention-derived dependencies and emphasizing attention-critical tokens during optimization, AGDO better matches training dynamics with the model’s internal reasoning structure. Experiments on mathematical and coding benchmarks demonstrate consistent improvements over state-of-the-art baselines, highlighting the effectiveness of attention-guided denoising and optimization for dLLMs.

\section*{Limitation}
Our experiments mainly focus on dLLMs with full attention, utilizing its properties in dLLMs to optimize algorithms. Therefore, we do not discuss the performance on block attention-based dLLMs. In future work, we plan to investigate the characteristics of block attention in dLLMs and design tailored algorithmic improvements.
\section*{Acknowledgments}
This work was partially supported by the National Natural Science Foundation of China No. 92470205 and Beijing Major Science and Technology Project under Contract No. Z251100008425002. 

\bibliography{custom}

@article{li2022diffusion,
  title={Diffusion-lm improves controllable text generation},
  author={Li, Xiang and Thickstun, John and Gulrajani, Ishaan and Liang, Percy S and Hashimoto, Tatsunori B},
  journal={Advances in neural information processing systems},
  volume={35},
  pages={4328--4343},
  year={2022}
}

@article{gong2022diffuseq,
  title={Diffuseq: Sequence to sequence text generation with diffusion models},
  author={Gong, Shansan and Li, Mukai and Feng, Jiangtao and Wu, Zhiyong and Kong, LingPeng},
  journal={arXiv preprint arXiv:2210.08933},
  year={2022}
}

@article{ye2025dream,
  title={Dream 7b: Diffusion large language models},
  author={Ye, Jiacheng and Xie, Zhihui and Zheng, Lin and Gao, Jiahui and Wu, Zirui and Jiang, Xin and Li, Zhenguo and Kong, Lingpeng},
  journal={arXiv preprint arXiv:2508.15487},
  year={2025}
}

@article{nie2025large,
  title={Large language diffusion models},
  author={Nie, Shen and Zhu, Fengqi and You, Zebin and Zhang, Xiaolu and Ou, Jingyang and Hu, Jun and Zhou, Jun and Lin, Yankai and Wen, Ji-Rong and Li, Chongxuan},
  journal={arXiv preprint arXiv:2502.09992},
  year={2025}
}

@article{cheng2025sdar,
  title={SDAR: A Synergistic Diffusion-AutoRegression Paradigm for Scalable Sequence Generation},
  author={Cheng, Shuang and Bian, Yihan and Liu, Dawei and Zhang, Linfeng and Yao, Qian and Tian, Zhongbo and Wang, Wenhai and Guo, Qipeng and Chen, Kai and Qi, Biqing and others},
  journal={arXiv preprint arXiv:2510.06303},
  year={2025}
}

@article{gong2024scaling,
  title={Scaling diffusion language models via adaptation from autoregressive models},
  author={Gong, Shansan and Agarwal, Shivam and Zhang, Yizhe and Ye, Jiacheng and Zheng, Lin and Li, Mukai and An, Chenxin and Zhao, Peilin and Bi, Wei and Han, Jiawei and others},
  journal={arXiv preprint arXiv:2410.17891},
  year={2024}
}

@article{li2025survey,
  title={A survey on diffusion language models},
  author={Li, Tianyi and Chen, Mingda and Guo, Bowei and Shen, Zhiqiang},
  journal={arXiv preprint arXiv:2508.10875},
  year={2025}
}

@article{dubey2024llama,
  title={The llama 3 herd of models},
  author={Dubey, Abhimanyu and Jauhri, Abhinav and Pandey, Abhinav and Kadian, Abhishek and Al-Dahle, Ahmad and Letman, Aiesha and Mathur, Akhil and Schelten, Alan and Yang, Amy and Fan, Angela and others},
  journal={arXiv e-prints},
  pages={arXiv--2407},
  year={2024}
}

@article{wu2025fast,
  title={Fast-dllm: Training-free acceleration of diffusion llm by enabling kv cache and parallel decoding},
  author={Wu, Chengyue and Zhang, Hao and Xue, Shuchen and Liu, Zhijian and Diao, Shizhe and Zhu, Ligeng and Luo, Ping and Han, Song and Xie, Enze},
  journal={arXiv preprint arXiv:2505.22618},
  year={2025}
}

@article{gao2025self,
  title={Self Speculative Decoding for Diffusion Large Language Models},
  author={Gao, Yifeng and Ji, Ziang and Wang, Yuxuan and Qi, Biqing and Xu, Hanlin and Zhang, Linfeng},
  journal={arXiv preprint arXiv:2510.04147},
  year={2025}
}

@article{liu2025dllm,
  title={dllm-cache: Accelerating diffusion large language models with adaptive caching},
  author={Liu, Zhiyuan and Yang, Yicun and Zhang, Yaojie and Chen, Junjie and Zou, Chang and Wei, Qingyuan and Wang, Shaobo and Zhang, Linfeng},
  journal={arXiv preprint arXiv:2506.06295},
  year={2025}
}

@article{bao2025learning,
  title={Learning to Parallel: Accelerating Diffusion Large Language Models via Learnable Parallel Decoding},
  author={Bao, Wenrui and Chen, Zhiben and Xu, Dan and Shang, Yuzhang},
  journal={arXiv preprint arXiv:2509.25188},
  year={2025}
}

@article{li2025adaptive,
  title={Adaptive Classifier-Free Guidance via Dynamic Low-Confidence Masking},
  author={Li, Pengxiang and Yan, Shilin and Tsai, Joey and Zhang, Renrui and An, Ruichuan and Guo, Ziyu and Gao, Xiaowei},
  journal={arXiv preprint arXiv:2505.20199},
  year={2025}
}

@misc{denoisingbeyond,
      title={Beyond Fixed: Training-Free Variable-Length Denoising for Diffusion Large Language Models}, 
      author={Jinsong Li and Xiaoyi Dong and Yuhang Zang and Yuhang Cao and Jiaqi Wang and Dahua Lin},
      year={2025},
      eprint={2508.00819},
      archivePrefix={arXiv},
}

@misc{khannAGDO025mercury,
      title={Mercury: Ultra-Fast Language Models Based on Diffusion}, 
      author={Inception Labs and Samar Khanna and Siddhant Kharbanda and Shufan Li and Harshit Varma and Eric Wang and Sawyer Birnbaum and Ziyang Luo and Yanis Miraoui and Akash Palrecha and Stefano Ermon and Aditya Grover and Volodymyr Kuleshov},
      year={2025},
      eprint={2506.17298},
      archivePrefix={arXiv},
      primaryClass={cs.CL},
      url={https://arxiv.org/abs/2506.17298}, 
}

@article{gong2025diffucoder,
  title={DiffuCoder: Understanding and Improving Masked Diffusion Models for Code Generation},
  author={Gong, Shansan and Zhang, Ruixiang and Zheng, Huangjie and Gu, Jiatao and Jaitly, Navdeep and Kong, Lingpeng and Zhang, Yizhe},
  journal={arXiv preprint arXiv:2506.20639},
  year={2025}
}

@article{tang2025wd1,
  title={wd1: Weighted policy optimization for reasoning in diffusion language models},
  author={Tang, Xiaohang and Dolga, Rares and Yoon, Sangwoong and Bogunovic, Ilija},
  journal={arXiv preprint arXiv:2507.08838},
  year={2025}
}

@article{wang2025spg,
  title={Spg: Sandwiched policy gradient for masked diffusion language models},
  author={Wang, Chengyu and Rashidinejad, Paria and Su, DiJia and Jiang, Song and Wang, Sid and Zhao, Siyan and Zhou, Cai and Shen, Shannon Zejiang and Chen, Feiyu and Jaakkola, Tommi and others},
  journal={arXiv preprint arXiv:2510.09541},
  year={2025}
}

@article{zhao2025d1,
  title={d1: Scaling reasoning in diffusion large language models via reinforcement learning},
  author={Zhao, Siyan and Gupta, Devaansh and Zheng, Qinqing and Grover, Aditya},
  journal={arXiv preprint arXiv:2504.12216},
  year={2025}
}

@article{zhu2025llada,
  title={LLaDA 1.5: Variance-Reduced Preference Optimization for Large Language Diffusion Models},
  author={Zhu, Fengqi and Wang, Rongzhen and Nie, Shen and Zhang, Xiaolu and Wu, Chunwei and Hu, Jun and Zhou, Jun and Chen, Jianfei and Lin, Yankai and Wen, Ji-Rong and others},
  journal={arXiv preprint arXiv:2505.19223},
  year={2025}
}

@article{shao2024deepseekmath,
  author       = {Zhihong Shao and
                  Peiyi Wang and
                  Qihao Zhu and
                  Runxin Xu and
                  Junxiao Song and
                  Mingchuan Zhang and
                  Y. K. Li and
                  Y. Wu and
                  Daya Guo},
  title        = {DeepSeekMath: Pushing the Limits of Mathematical Reasoning in Open
                  Language Models},
  journal      = {CoRR},
  volume       = {abs/2402.03300},
  year         = {2024},
  url          = {https://doi.org/10.48550/arXiv.2402.03300},
  doi          = {10.48550/ARXIV.2402.03300},
  eprinttype    = {arXiv},
  eprint       = {2402.03300},
  bibsource    = {dblp computer science bibliography, https://dblp.org}
}

@article{schulman2017proximal,
  title={Proximal policy optimization algorithms},
  author={Schulman, John and Wolski, Filip and Dhariwal, Prafulla and Radford, Alec and Klimov, Oleg},
  journal={arXiv preprint arXiv:1707.06347},
  year={2017}
}

@article{yang2025qwen3,
  title={Qwen3 technical report},
  author={Yang, An and Li, Anfeng and Yang, Baosong and Zhang, Beichen and Hui, Binyuan and Zheng, Bo and Yu, Bowen and Gao, Chang and Huang, Chengen and Lv, Chenxu and others},
  journal={arXiv preprint arXiv:2505.09388},
  year={2025}
}

@article{cobbe2021training,
  title={Training verifiers to solve math word problems},
  author={Cobbe, Karl and Kosaraju, Vineet and Bavarian, Mohammad and Chen, Mark and Jun, Heewoo and Kaiser, Lukasz and Plappert, Matthias and Tworek, Jerry and Hilton, Jacob and Nakano, Reiichiro and others},
  journal={arXiv preprint arXiv:2110.14168},
  year={2021}
}

@article{hendrycks2020measuring,
  title={Measuring massive multitask language understanding},
  author={Hendrycks, Dan and Burns, Collin and Basart, Steven and Zou, Andy and Mazeika, Mantas and Song, Dawn and Steinhardt, Jacob},
  journal={arXiv preprint arXiv:2009.03300},
  year={2020}
}

@article{lewkowycz2022solving,
  title={Solving quantitative reasoning problems with language models},
  author={Lewkowycz, Aitor and Andreassen, Anders and Dohan, David and Dyer, Ethan and Michalewski, Henryk and Ramasesh, Vinay and Slone, Ambrose and Anil, Cem and Schlag, Imanol and Gutman-Solo, Theo and others},
  journal={Advances in neural information processing systems},
  volume={35},
  pages={3843--3857},
  year={2022}
}

@article{white2024livebench,
  title={Livebench: A challenging, contamination-free llm benchmark},
  author={White, Colin and Dooley, Samuel and Roberts, Manley and Pal, Arka and Feuer, Ben and Jain, Siddhartha and Shwartz-Ziv, Ravid and Jain, Neel and Saifullah, Khalid and Naidu, Siddartha and others},
  journal={arXiv preprint arXiv:2406.19314},
  volume={4},
  year={2024}
}

@article{jain2024livecodebench,
  title={Livecodebench: Holistic and contamination free evaluation of large language models for code},
  author={Jain, Naman and Han, King and Gu, Alex and Li, Wen-Ding and Yan, Fanjia and Zhang, Tianjun and Wang, Sida and Solar-Lezama, Armando and Sen, Koushik and Stoica, Ion},
  journal={arXiv preprint arXiv:2403.07974},
  year={2024}
}

@article{wang2025revolutionizing,
  title={Revolutionizing reinforcement learning framework for diffusion large language models},
  author={Wang, Yinjie and Yang, Ling and Li, Bowen and Tian, Ye and Shen, Ke and Wang, Mengdi},
  journal={arXiv preprint arXiv:2509.06949},
  year={2025}
}

@article{team2024qwen2,
  title={Qwen2 technical report},
  author={Team, Qwen and others},
  journal={arXiv preprint arXiv:2407.10671},
  volume={2},
  number={3},
  year={2024}
}

@article{li2022competition,
  title={Competition-level code generation with alphacode},
  author={Li, Yujia and Choi, David and Chung, Junyoung and Kushman, Nate and Schrittwieser, Julian and Leblond, R{\'e}mi and Eccles, Tom and Keeling, James and Gimeno, Felix and Dal Lago, Agustin and others},
  journal={Science},
  volume={378},
  number={6624},
  pages={1092--1097},
  year={2022},
  publisher={American Association for the Advancement of Science}
}

@article{jaghouar2024intellect,
  title={Intellect-1 technical report},
  author={Jaghouar, Sami and Ong, Jack Min and Basra, Manveer and Obeid, Fares and Straube, Jannik and Keiblinger, Michael and Bakouch, Elie and Atkins, Lucas and Panahi, Maziyar and Goddard, Charles and others},
  journal={arXiv preprint arXiv:2412.01152},
  year={2024}
}

@article{llama3modelcard,

title={Llama 3 Model Card},

author={AI@Meta},

year={2024},

url = {https://github.com/meta-llama/llama3/blob/main/MODEL_CARD.md}

}

@article{sun2025blockwise,
  title={Blockwise sft for diffusion language models: Reconciling bidirectional attention and autoregressive decoding},
  author={Sun, Bowen and Cai, Yujun and Yang, Ming-Hsuan and Wang, Yiwei},
  journal={arXiv preprint arXiv:2508.19529},
  year={2025}
}

@article{song2025sparse,
  title={Sparse-dllm: Accelerating diffusion llms with dynamic cache eviction},
  author={Song, Yuerong and Liu, Xiaoran and Li, Ruixiao and Liu, Zhigeng and Huang, Zengfeng and Guo, Qipeng and He, Ziwei and Qiu, Xipeng},
  journal={arXiv preprint arXiv:2508.02558},
  year={2025}
}

@article{xiao2023efficient,
  title={Efficient streaming language models with attention sinks},
  author={Xiao, Guangxuan and Tian, Yuandong and Chen, Beidi and Han, Song and Lewis, Mike},
  journal={arXiv preprint arXiv:2309.17453},
  year={2023}
}

@inproceedings{hsieh2024found,
  title={Found in the middle: Calibrating positional attention bias improves long context utilization},
  author={Hsieh, Cheng-Yu and Chuang, Yung-Sung and Li, Chun-Liang and Wang, Zifeng and Le, Long and Kumar, Abhishek and Glass, James and Ratner, Alexander and Lee, Chen-Yu and Krishna, Ranjay and others},
  booktitle={Findings of the Association for Computational Linguistics: ACL 2024},
  pages={14982--14995},
  year={2024}
}

@article{cheng2025deer,
  title={DEER: Draft with Diffusion, Verify with Autoregressive Models},
  author={Cheng, Zicong and Yang, Guo-Wei and Li, Jia and Deng, Zhijie and Guo, Meng-Hao and Hu, Shi-Min},
  journal={arXiv preprint arXiv:2512.15176},
  year={2025}
}

@article{li2025attention,
  title={Attention illuminates llm reasoning: The preplan-and-anchor rhythm enables fine-grained policy optimization},
  author={Li, Yang and Dong, Zhichen and Sun, Yuhan and Wang, Weixun and Xiong, Shaopan and Luo, Yijia and Liu, Jiashun and Lu, Han and Wang, Jiamang and Su, Wenbo and others},
  journal={arXiv preprint arXiv:2510.13554},
  year={2025}
}

@article{lin2024critical,
  title={Critical Tokens Matter: Token-Level Contrastive Estimation Enhances LLM's Reasoning Capability},
  author={Lin, Zicheng and Liang, Tian and Xu, Jiahao and Lin, Qiuzhi and Wang, Xing and Luo, Ruilin and Shi, Chufan and Li, Siheng and Yang, Yujiu and Tu, Zhaopeng},
  journal={arXiv preprint arXiv:2411.19943},
  year={2024}
}

@inproceedings{clark2019does,
  title={What does BERT look at? an analysis of BERT’s attention},
  author={Clark, Kevin and Khandelwal, Urvashi and Levy, Omer and Manning, Christopher D},
  booktitle={Proceedings of the 2019 ACL workshop BlackboxNLP: analyzing and interpreting neural networks for NLP},
  pages={276--286},
  year={2019}
}

@inproceedings{zellers2019hellaswag,
  title={Hellaswag: Can a machine really finish your sentence?},
  author={Zellers, Rowan and Holtzman, Ari and Bisk, Yonatan and Farhadi, Ali and Choi, Yejin},
  booktitle={Proceedings of the 57th annual meeting of the association for computational linguistics},
  pages={4791--4800},
  year={2019}
}

@inproceedings{talmor2019commonsenseqa,
  title={Commonsenseqa: A question answering challenge targeting commonsense knowledge},
  author={Talmor, Alon and Herzig, Jonathan and Lourie, Nicholas and Berant, Jonathan},
  booktitle={Proceedings of the 2019 Conference of the North American Chapter of the Association for Computational Linguistics: Human Language Technologies, Volume 1 (Long and Short Papers)},
  pages={4149--4158},
  year={2019}
}

\appendix

\section{Attention Analysis on LLaDA}
\label{app_llada}
To assess the generalizability of our findings beyond the Dream model, we extend our empirical analysis to LLaDA-8B-Instruct~\citep{nie2025large}.

\begin{figure}[h!]
  \centering
  \begin{subfigure}[b]{0.49\linewidth}
    \centering
    \includegraphics[width=\linewidth]{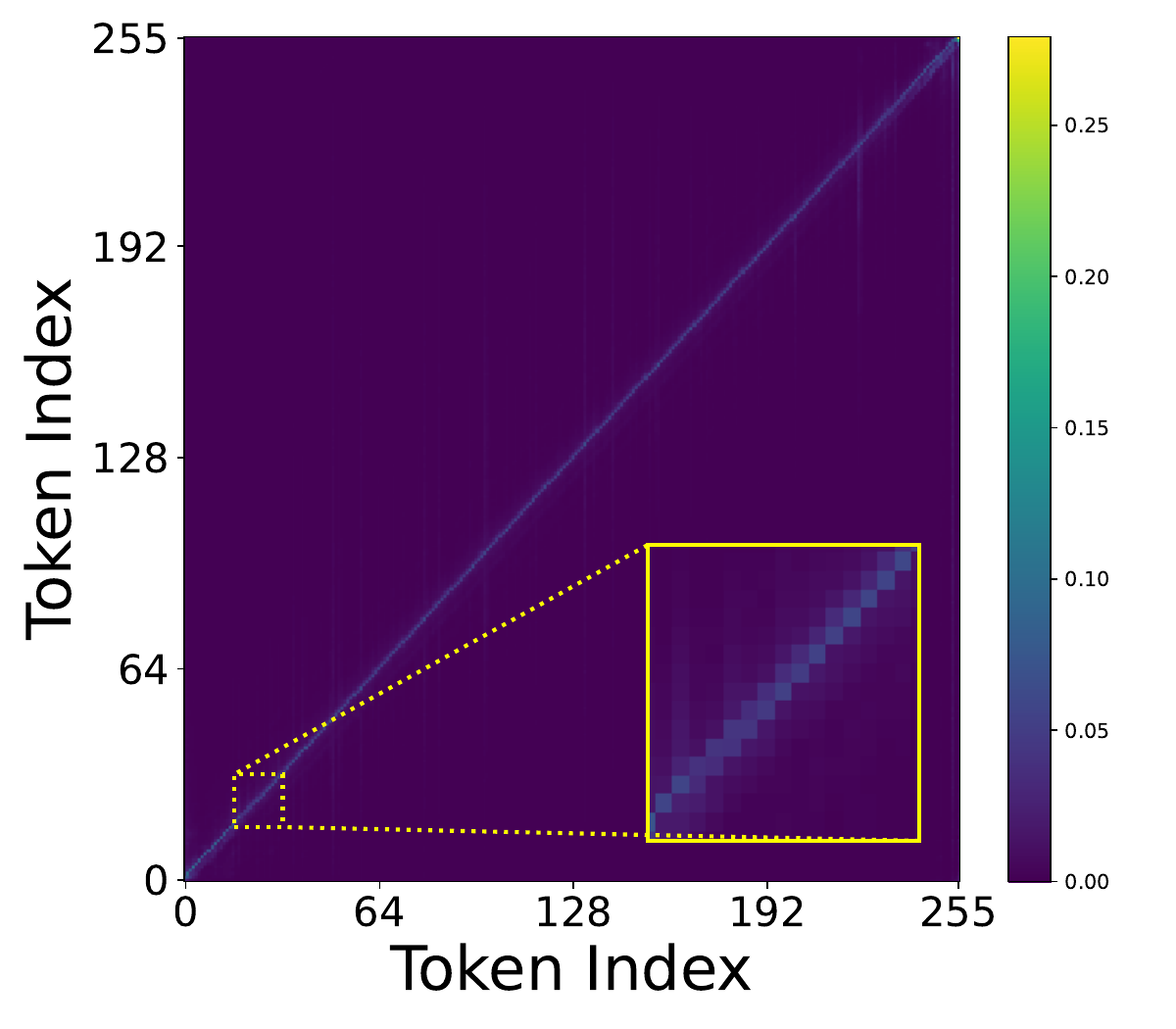}
    \caption{Step 136 attention map.}
    \label{fig:llada_map_136}
  \end{subfigure}
  \hfill
  \begin{subfigure}[b]{0.49\linewidth}
    \centering
    \includegraphics[width=\linewidth]{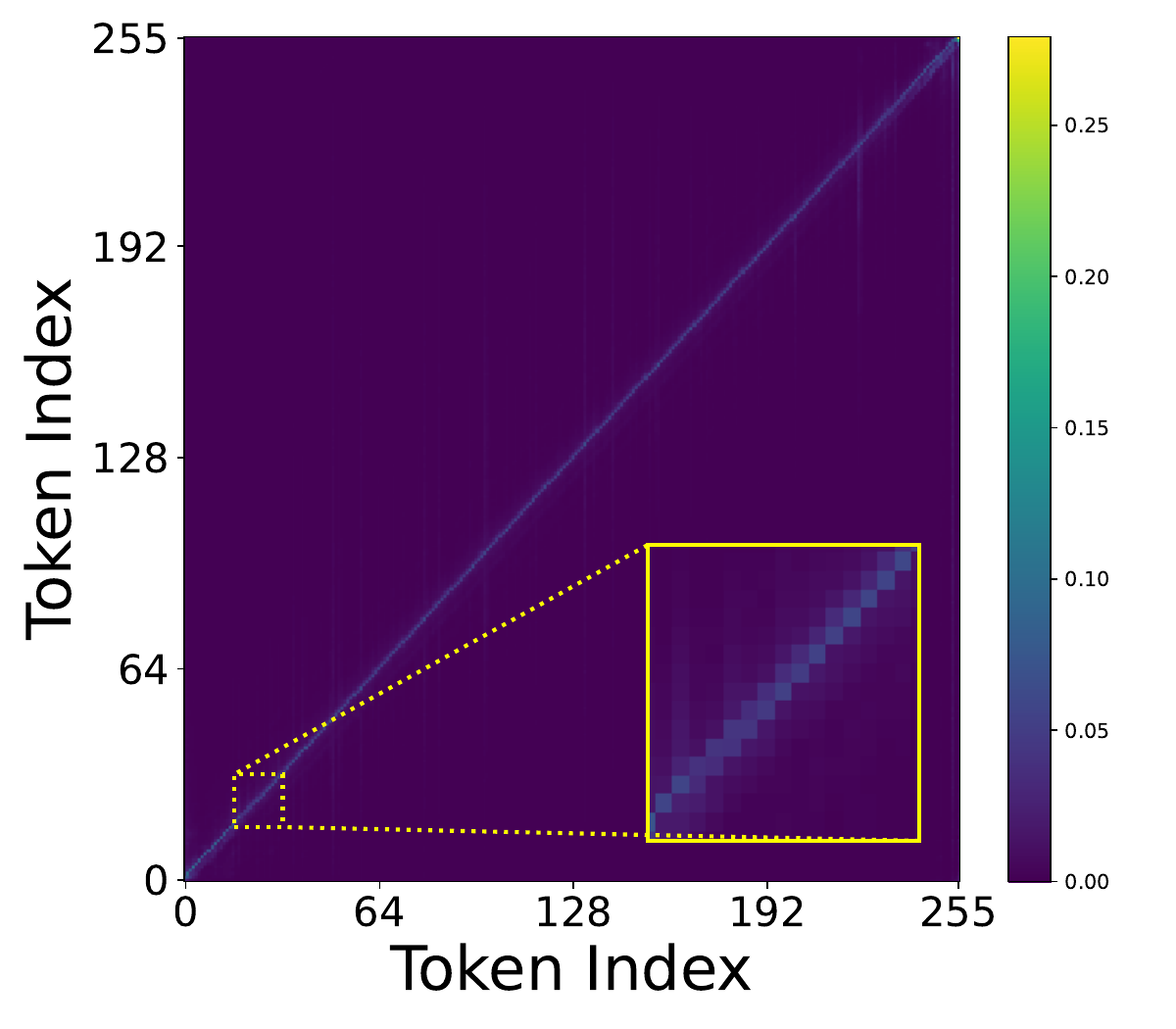}
    \caption{Step 237 attention map.}
    \label{fig:llada_map_237}
  \end{subfigure}
  
  \vspace{1em} 
  
  \begin{subfigure}[b]{0.49\linewidth}
    \centering
    \includegraphics[width=\linewidth]{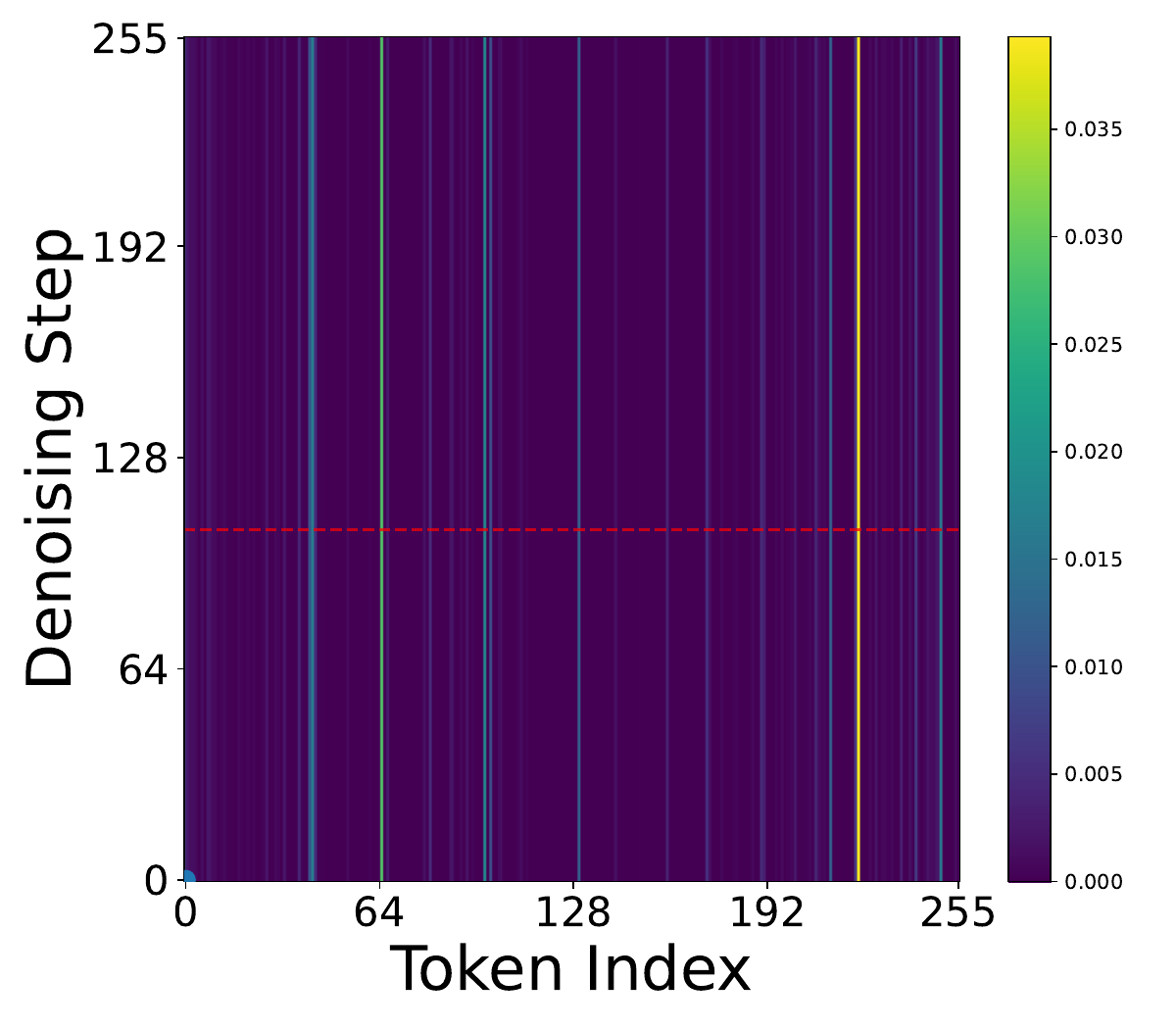}
    \caption{Token 81 attention change.}
    \label{fig:llada_token_81}
  \end{subfigure}
  \hfill
  \begin{subfigure}[b]{0.49\linewidth}
    \centering
    \includegraphics[width=\linewidth]{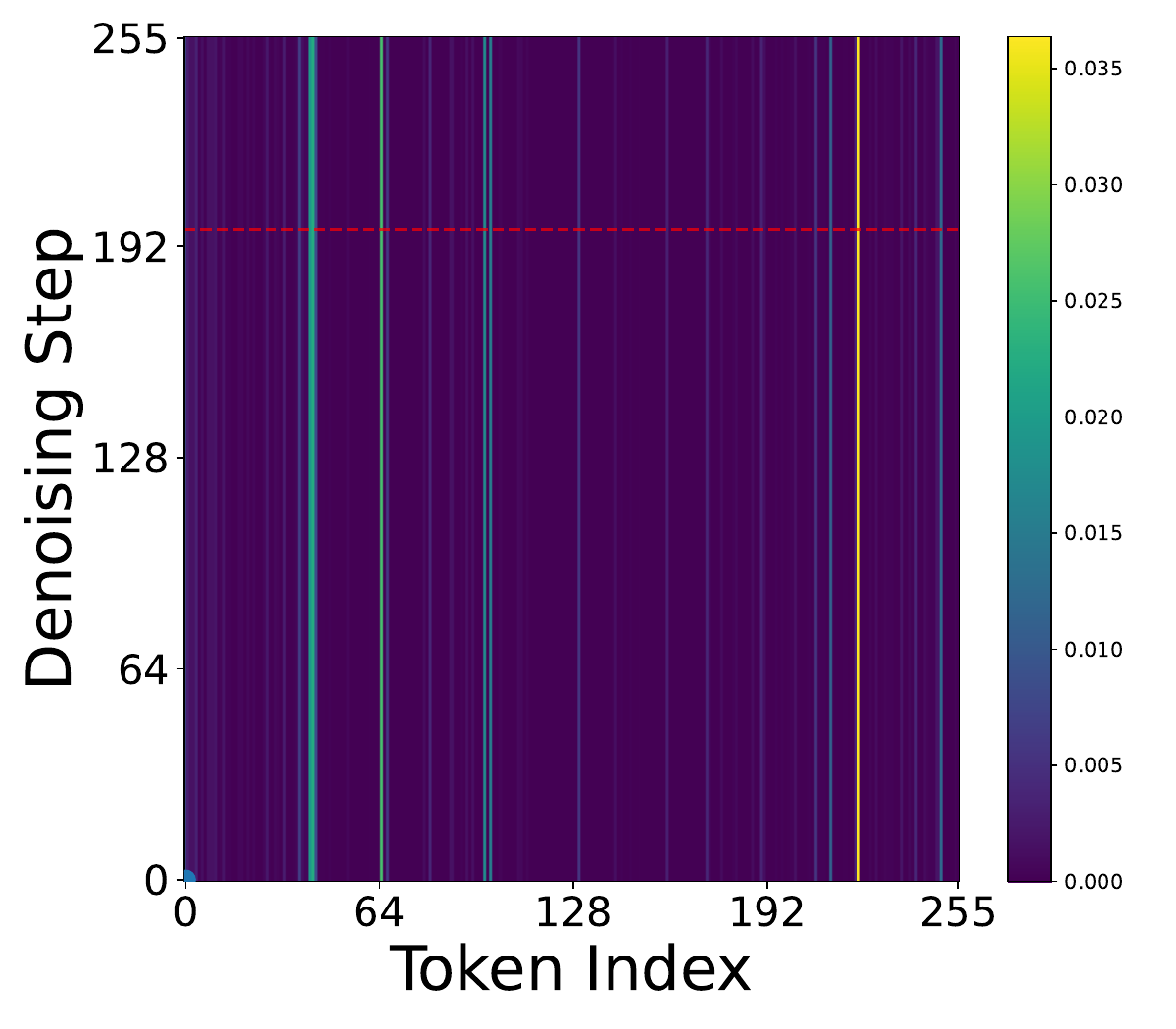}
    \caption{Token 20 attention change.}
    \label{fig:llada_token_200}
  \end{subfigure}

  \caption{Analysis of attention patterns on LLaDA. (a) and (b) visualize the attention distributions among tokens at two randomly selected denoising steps. (c) and (d) illustrate the temporal attention dynamics of specific tokens towards others throughout the denoising process.}
  \label{fig:llada_combined_results}
\end{figure}

As illustrated in Figure~\ref{fig:llada_combined_results}, LLaDA displays attention patterns comparable to those observed in the Dream model, notably exhibiting horizontal sparsity and vertical consistency. Furthermore, the results in Table~\ref{tab:accuracy_comparison-llada} suggest that aligning the decoding order with intrinsic attention dependencies could serve as a generalized principle for enhancing generation quality in dLLMs.

\begin{table}[!h]
    \centering
    \resizebox{0.8\linewidth}{!}{ 
        \setlength{\tabcolsep}{12pt}
        \begin{tabular}{lccc}
            \toprule
            \textbf{Strategy} & \textbf{128} & \textbf{256} & \textbf{512} \\
            \midrule
            Max-Prob & 19.9 & 23.8 & 16.1 \\ 
            Max-$S$    & 28.5 & 30.3 & 32.7 \\
            \bottomrule
        \end{tabular}
    }
     \caption{Comparison of decoding accuracy on MATH500 using static sampling across different predefined sequence lengths on LLaDA. For the Max-$S$ strategy, the probability threshold is set to 0.}
     \label{tab:accuracy_comparison-llada}
\end{table}
\section{Detailed Algorithms}
\label{app:algorithms}

Here we provide the pseudocode for the three core components of the AGDO framework.
\subsection{Attention-Guided Denoising Order}
\label{app:AGDOtg}
\begin{algorithm}[H]
\small
\caption{Attention-Guided Denoising Order}
\label{alg:trajectory}
\begin{algorithmic}[1]
    \State \textbf{Input:} Sequence $X$; Model $\pi_\theta$; Step size $n$; Block size $m$
    \State \textbf{Output:} Denoising Traces $T \in \mathbb{Z}^{|X|}$
    
    \State \textbf{Init:} $\mathcal{U} \leftarrow \text{Prompt}(X); \ \mathcal{M} \leftarrow \text{Masked}(X); \ T \leftarrow \mathbf{0}; \ \text{rank} \leftarrow 1$
    \State $A \leftarrow \text{Forward}(X, \pi_\theta)$ 
    
    \While{$\mathcal{M} \neq \emptyset$}
        \State $\forall i \in \mathcal{M}: S_i \leftarrow \sum_{k \in \mathcal{U}} \text{MeanHead}(A_{i, k})$
        \State $\mathcal{K} \leftarrow \text{TopK}(\{S_i\}, n)$
        \State $\forall k \in \mathcal{K}: T_k \leftarrow \text{rank}; \quad \mathcal{U} \leftarrow \mathcal{U} \cup \mathcal{K}; \quad \mathcal{M} \leftarrow \mathcal{M} \setminus \mathcal{K}$
        \State $\text{rank} \leftarrow \text{rank} + 1$
    \EndWhile
    
    \State \textbf{Constant} $\mathcal{C} \leftarrow |X|$
    \For{$idx = 0 \dots |X|-1$}
        \State $\text{batch\_num} \leftarrow \lfloor idx / m \rfloor$
        \State $T_{idx} \leftarrow T_{idx} + (\text{batch\_num} + 1) \cdot \mathcal{C}$
    \EndFor
    
    \State \textbf{Return} $T$
\end{algorithmic}
\end{algorithm}

\subsection{AGDO-SFT}
\label{app:AGDOft}
\begin{algorithm}[H]
\small
\caption{AGDO-SFT}
\label{alg:AGDOft}
\begin{algorithmic}[1]
    \State \textbf{Input:} Dataset $\mathcal{D}$; Model $\pi_\theta$; Rate $\eta$; Coeff $\gamma$
    
    \Statex
    \For{batch $B \in \mathcal{D}$}
        \State $\mathcal{L} \leftarrow 0$
        
        \For{$X \in B$}
            \State $T \leftarrow \text{Alg.\ref{alg:trajectory}}(X); \quad A \leftarrow \text{Forward}(X, \pi_\theta)$
            \State $\forall k: I_k \leftarrow \sum_{i} \text{MeanHead}(A_{i, k})$ \Comment{Eq.~\ref{equ:influence}}
            
            \For{$t = 0 \dots \max(T)$}
                \State $\mathcal{K}_t \leftarrow \{k \mid D_k = t\}$
                
                \State $\mathcal{M}_{sub} \leftarrow \text{RandomMask}(\mathcal{K}_t)$
                
                \State $\mathcal{M}_{in} \leftarrow \{k \mid D_k > t\} \cup \mathcal{M}_{sub}$
                
                \State $W \leftarrow \{1 + \gamma I_k \mid k \in \mathcal{M}_{sub}\}$
                
                \State $\mathcal{L} \leftarrow \mathcal{L} + \text{WeightedNLL}(\pi_\theta(X_{\mathcal{M}_{in}}), \mathcal{M}_{sub}, W)$ \Comment{Eq.~\ref{eq:attn_received_score}}
            \EndFor
        \EndFor
        
        \State $\theta \leftarrow \theta - \eta \nabla_\theta \mathcal{L}$
    \EndFor
\end{algorithmic}
\end{algorithm}

\subsection{AGDO-RL}
\label{app:AGDOpo}
\begin{algorithm}[H]
\small
\caption{AGDO-RL}
\label{alg:AGDOpo}
\begin{algorithmic}[1]
    \State \textbf{Input:} Prompts $\mathcal{Q}$; Ref $\pi_{ref}$; Group $G$; Coeff $\delta$
    
    \Statex
    \For{iteration $1 \dots \text{MaxIter}$}
        \State $q \sim \mathcal{Q}$
        
        \State $\{o_1 \dots o_G\} \leftarrow \text{Gen}(q, \pi_\theta)$
        
        \State $R \leftarrow \text{Reward}(q, \{o_j\}); \quad \hat{A} \leftarrow \text{Norm}(R)$
        
        \State $\mathcal{L} \leftarrow 0$
        
        \For{$j = 1 \dots G$}
            \State $T \leftarrow \text{Alg.\ref{alg:trajectory}}([q, o_j], \pi_\theta); \quad A \leftarrow \text{Forward}([q, o_j], \pi_\theta)$
            
            \State $\forall k: I_k \leftarrow \sum_{i} \text{MeanHead}(A_{i, k})$ \Comment{Eq.~\ref{equ:influence}}
            \State $\forall k: \hat{A}'_k \leftarrow \hat{A}_j + \text{sign}(\hat{A}_j) \cdot \delta \cdot I_k$ \Comment{Eq.~\ref{equ:adv_new}}
            
            \For{$t = 0 \dots \max(T)$}
                \State $\mathcal{K}_t \leftarrow \{k \mid T_k = t\}$
                
                \State $\mathcal{M}_{sub} \leftarrow \text{RandomMask}(\mathcal{K}_t)$
                
                \State $\mathcal{M}_{in} \leftarrow \{k \mid T_k > t\} \cup \mathcal{M}_{sub}$
                
                \State $\pi_{curr} \leftarrow \pi_\theta(o_j[\mathcal{M}_{sub}] \mid [q, o_j]_{\mathcal{M}_{in}})$
                \State $\pi_{old} \leftarrow \pi_{ref}(o_j[\mathcal{M}_{sub}] \mid [q, o_j]_{\mathcal{M}_{in}})$
                
                \State $\mathcal{L} \leftarrow \mathcal{L} + \text{GRPO\_Obj}(\pi_{curr}, \pi_{old}, \hat{A}'_{\mathcal{M}_{sub}})$ \Comment{Eq.~\ref{equ:grpo_dllm}}
            \EndFor
        \EndFor
        
        \State $\theta \leftarrow \theta - \eta \nabla_\theta \mathcal{L}$
    \EndFor
\end{algorithmic}
\end{algorithm}
\section{Implementation Details}
\label{app_imp}
All experiments are conducted on NVIDIA H20 GPUs.
\textbf{Supervised Fine-Tuning (SFT):}
For mathematical reasoning, we use 2,000 training samples from \citet{wang2025revolutionizing}. For code generation, we distill 2,000 samples from CodeContest~\citep{li2022competition} using Qwen2.5-32B-Instruct~\citep{team2024qwen2}. Models are trained for one epoch with a learning rate of $1\times10^{-6}$ and a total batch size of 128. We set \texttt{post\_num} to 16 and $\gamma$ to 100. \textbf{Reinforcement Learning (RL):}
We adopt the training framework from \citet{wang2025revolutionizing}. The training data consists of the training splits of GSM8K and MATH for mathematical reasoning, and PrimeIntellect~\citep{jaghouar2024intellect} for code generation. For coding tasks, the reward is defined as the fraction of unit tests passed. During rollout, we sample 32 prompts per step and generate 8 responses for each prompt. We employ a static denoising strategy with a temperature of 1.0, unmasking one token per denoising step, and limit the maximum sequence length to 1,024 tokens. The total batch size is set to 512, with a learning rate of $1\times10^{-6}$. We set $\delta$ to 1. The \texttt{post\_num} is set to 16 for mathematical tasks and 0 for code generation.
\end{document}